\documentclass[letterpaper]{article}
\usepackage{uai2020}
\usepackage[margin=1in]{geometry}

\usepackage{times}

\usepackage[authoryear,sort,comma,numbers]{natbib}

\usepackage[utf8x]{inputenc} 
\usepackage{hyperref}       
\usepackage{url}            
\usepackage{booktabs}       
\usepackage{nicefrac}       
\usepackage{microtype}      

\usepackage{graphicx}
\usepackage{amsmath}
\usepackage{amssymb}
\usepackage{bbold}
\usepackage{bm}
\usepackage{cases}
\usepackage{color}
\usepackage{amsthm}
\newtheorem{theorem}{Theorem}
\newtheorem*{theorem*}{Theorem}

\usepackage{pgf}
\usepackage{tikz}
\usepackage{pgfplots}
\usepackage{pgf-umlsd}
\usepackage{ifthen}
\usetikzlibrary{shapes, snakes, positioning, arrows,calc}
\pgfplotsset{compat=newest}
\usepgfplotslibrary{groupplots}
\usepgfplotslibrary{dateplot}
\usepackage{enumitem}

\usepackage{algorithm}
\usepackage{algpseudocode}
\usepackage{wrapfig}
\usepackage{dblfloatfix}

\usepackage{caption}
\usepackage{subcaption}

\DeclareCaptionFont{scriptsize}{\scriptsize}
\usepackage[compact]{titlesec}
\titlespacing\section{2pt}{2pt plus 0pt minus 2pt}{0pt plus 0pt minus 1pt}
\titlespacing\subsection{2pt}{2pt plus 0pt minus 2pt}{1pt plus 0pt minus 2pt}
\titlespacing\subsubsection{2pt}{2pt plus 0pt minus 2pt}{1pt plus 0pt minus 2pt}
\titlespacing\paragraph{0pt}{0pt plus 0pt minus 2pt}{7pt plus 2pt minus 0pt}
\def\clap#1{\hbox to 0pt{\hss#1\hss}}

\def\mathclap{\mathpalette\mathclapinternal}

\def\mathclapinternal#1#2{%
	\clap{$\mathsurround=0pt#1{#2}$}}

\makeatletter
\g@addto@macro\normalsize{%
	\setlength\abovedisplayskip{3pt}
	\setlength\belowdisplayskip{3pt}
	\setlength\abovedisplayshortskip{0pt}
	\setlength\belowdisplayshortskip{0pt}
}
\makeatother

\usepackage{etoolbox}
\usepackage{nicefrac}

\newcommand{\mbf}[1]{\mathbf{#1}}
\newcommand{\zbhat}{\mbf{\hat{z}}}
\newcommand{\zb}{\mbf{z}}
\newcommand{\nub}{\bm{\nu}}
\newcommand{\mub}{\bm{\mu}}

\newcommand{\lambdab}{\bm{\lambda}}
\newcommand{\lambdabhat}{\bm{\rho}}

\newcommand{\ub}{\mbf{u}}
\newcommand{\lb}{\mbf{l}}
\newcommand{\xb}{\mbf{x}}
\newcommand{\tb}{\mbf{t}}
\newcommand{\xbhat}{\mbf{\hat{x}}}
\newcommand{\mtt}{\mathtt}

\newcommand{\sig}[1]{\sigma\left(#1\right)}
\newcommand{\phiub}{\phi_{\ub_k, \lb_k}}
\newcommand{\psiub}{\psi_{\ub_k, \lb_k}}
\newcommand{\pib}{\bm{\pi}}
\newcommand{\ulslope}{\frac{\sig{\ub_k}-\sig{\lb_k}}{\ub_k - \lb_k}}
\newcommand{\tlslope}{\frac{\sig{\tb_k} - \sig{\lb_k}}{\tb_k - \lb_k}}
\newcommand{\auglag}{\mathcal{L}\left( \zbhat,\lambdabhat\right)}

\DeclareMathOperator*{\argmin}{\arg\!\min}

\newtoggle{appendix}
\newtoggle{experiments}
\newtoggle{longversion}

\toggletrue{appendix}
\togglefalse{longversion}

\newcommand\Tstrut{\rule{0pt}{2.6ex}}       
\newcommand\Bstrut{\rule[-0.9ex]{0pt}{0pt}} 
\newcommand{\TBstrut}{\Tstrut\Bstrut} 
\renewcommand{\bfseries}{\fontseries{b}\selectfont}
\newrobustcmd{\B}{\bfseries}

\usepackage{array}
\usepackage{multirow}
\usepackage{wrapfig}
\usepackage{subcaption}
\usepackage{tikz}
\usepackage{booktabs}
\usepackage{adjustbox}
\usetikzlibrary{shapes, positioning, arrows,calc}
\usepackage{siunitx}
\usepackage{etoolbox}
\usepackage{soul}

\title{Lagrangian Decomposition for Neural Network Verification}

\author{%
	Rudy Bunel\thanks{\hspace{1pt} These authors contributed equally to this work. \vspace{10pt}},\hspace{2pt} Alessandro De Palma\footnotemark[1],\hspace{3pt} Alban Desmaison \\
	University of Oxford\\
	$\{$\texttt{rudy}, \texttt{adepalma}$\}$\texttt{@robots.ox.ac.uk} 
	\AND
	Krishnamurthy (Dj) Dvijotham, Pushmeet Kohli \\
	DeepMind\\
	\And
	Philip H.S. Torr, M. Pawan Kumar \\
	University of Oxford\\
}

\begin{document}

\maketitle

\begin{abstract}
A fundamental component of neural network verification is the computation of bounds on the values their outputs can take.
Previous methods have either used off-the-shelf solvers, discarding the problem structure, or relaxed the problem even further, making the bounds unnecessarily loose.
We propose a novel approach based on Lagrangian Decomposition. Our formulation admits an efficient supergradient ascent algorithm, as well as an improved proximal algorithm. Both the algorithms offer three advantages: 
(i) they yield bounds that are provably at least as tight as previous dual algorithms relying on Lagrangian relaxations;
(ii) they are based on operations analogous to forward/backward pass of neural networks layers and are therefore easily parallelizable, amenable to GPU implementation and able to take advantage of the convolutional structure of problems; and 
(iii) they allow for anytime stopping while still providing valid bounds.
Empirically, we show that we obtain bounds comparable with off-the-shelf solvers in a fraction of their running time, and obtain tighter bounds in the same time as previous dual algorithms. This results in an overall speed-up when employing the bounds for formal verification.
Code for our algorithms is available at \url{https://github.com/oval-group/decomposition-plnn-bounds}.
\end{abstract}

\section{INTRODUCTION}
As deep learning powered systems become more and more common, the lack of
robustness of neural networks and their reputation for being ``Black Boxes'' is
increasingly worrisome. 
In order to deploy them in critical scenarios where safety and robustness would be a prerequisite, we need to invent techniques that can prove formal guarantees for neural network behaviour.
A particularly desirable property is resistance to adversarial
examples~\citep{Goodfellow2015,Szegedy2014}: perturbations maliciously crafted
with the intent of fooling even extremely well performing models. After
several defenses were proposed and subsequently broken \citep{Athalye2018, Uesato2018}, some progress has been made in being able to formally verify whether there exist any adversarial examples in the
neighbourhood of a data point~\citep{Tjeng2019, Wong2018}.

Verification algorithms fall into three categories: unsound (some false properties are proven false), incomplete (some true properties are proven true), and complete (all properties are correctly verified as either true or false).
A critical component of the verification systems developed so far is the computation of lower and upper bounds on the output of neural
networks when their inputs are constrained to lie in a bounded set.
In incomplete verification, by deriving bounds
on the changes of the prediction vector under restricted perturbations, it is possible to identify safe regions of the input space. These
results allow the rigorous comparison of adversarial defenses and prevent making
overconfident statements about their efficacy~\citep{Wong2018}. 
In complete verification, bounds can also be used as essential subroutines of Branch and Bound complete
verifiers~\citep{Bunel2018}.
Finally, bounds might also
be used as a training signal to guide the network towards greater
robustness and more verifiability~\citep{Gowal2018b,Mirman2018,Wong2018}.

Most previous algorithms for computing bounds are either computationally
expensive~\citep{Ehlers2017} or sacrifice a lot of tightness in order to
scale~\citep{Gowal2018b,Mirman2018,Wong2018}. In this work, we design novel customised relaxations and their corresponding solvers for obtaining bounds on neural networks. Our approach offers
the following advantages:

\begin{itemize}[nosep,leftmargin=1em,labelwidth=*,align=left]
	\item While previous approaches to neural network bounds~\citep{Dvijotham2018} are based on Lagrangian relaxations, we derive a new family of optimization problems for neural network bounds through {\bf Lagrangian Decomposition}, which in general yields duals at least as strong as those obtained through Lagrangian relaxation \citep{Guignard1987}. We in fact prove that, in the context of ReLU networks, for any dual solution from the approach by \citet{Dvijotham2018}, the bounds output by our dual are as least as tight.
	We demonstrate empirically that this derivation computes tighter bounds in the same time when using supergradient methods. We further improve on the performance by devising a proximal solver for the problem, which decomposes the task into a series of strongly convex subproblems. For each, we use an iterative method for which we derive
	optimal step sizes.
	\item Both the supergradient and the proximal method operate through linear operations similar to those
	used during network forward/backward passes.  As a consequence, we can {\bf leverage the
		convolutional structure} when necessary, while standard solvers are often
	restricted to treating it as a general linear operation.
	Moreover, both methods are {\bf easily parallelizable}:
	when computing bounds on the activations at layer $k$, we need two solve two problems for each hidden unit of the network (for the upper and lower bounds). These can all be solved in parallel.
	In complete verification, we need to compute bounds for several different problem domains at once: we solve these problems in parallel as well.
	Our GPU implementation thus allows us to solve several hundreds of linear programs at once on a single GPU, a level of parallelism that would be hard to match on CPU-based systems.

	\item Most standard linear programming based relaxations~\citep{Ehlers2017} will only
	return a valid bound if the problem is solved to optimality. Others, like the dual simplex method employed by off-the-shelf solvers~\citep{gurobi-custom} have a very high cost per iteration and will not yield tight bounds without incurring significant computational costs.
	Both methods described in this paper are {\bf anytime} (terminating it before convergence still provides a valid bound), and can be interrupted at very small granularity. This is useful in the
	context of a subroutine for complete verifiers, as this enables the user to
	choose an appropriate speed versus accuracy trade-off. It also offers great versatility as an incomplete verification method.
\end{itemize}

\section{RELATED WORKS}
Bound computations are mainly used for formal verification methods. Some
methods are complete~\citep{Cheng2017,Ehlers2017,Katz2017,Tjeng2019,Xiang2017},
always returning a verdict for each problem instances. Others are incomplete, based on relaxations of the verification problem. They trade speed for
completeness: while they cannot verify properties for all problem instances, they
scale significantly better. Two main types of bounds have been proposed: on the one hand,
some approaches~\citep{Ehlers2017,Salman2019} rely on off-the-shelf solvers to
solve accurate relaxations such as PLANET~\citep{Ehlers2017}, which is the best known
linear-sized approximation of the problem. On the other hand, as PLANET and other
more complex relaxations do not have closed form solutions, some researchers have also
proposed easier to solve, looser\iftoggle{longversion}{formulations. For
	example, \citet{Wong2018} considered the Lagrangian dual of the problem and
	used a specific choice of dual variables to derive a bound. \citet{Weng2018}
	achieved an equivalent relaxation by not taking the convex hull of the ReLU,
	but a larger overapproximation, leading to a problem with a closed form
	solution. Others relaxed the problem even further in order to obtain faster
	solutions, either by propagating intervals~\citep{Gowal2018b}, or through
	abstract interpretation~\citep{Mirman2018}. }{
	formulations~\citep{Gowal2018b,Mirman2018,Weng2018,Wong2018}.} Explicitly or
implicitly, these are all equivalent to propagating a convex domain through the
network to overapproximate the set of reachable values. Our approach consists in
tackling a relaxation equivalent to the PLANET one (although generalised beyond
ReLU), by designing a custom solver that achieves faster performance without
sacrificing tightness.
Some potentially tighter convex relaxations exist but involve a
quadratic number of variables, such as the semi-definite programming method of
\citet{Raghunathan2018} . Better relaxations obtained from relaxing strong Mixed
Integer Programming formulations~\citep{Anderson2019} have a quadratic
number of variables or a potentially exponential number of constraints. We do not
address them here.

A closely related approach to ours is the work of \citet{Dvijotham2018}. Both their
method and ours are anytime and operate on similar duals. 
While their dual is based on the Lagrangian relaxation of the non-convex problem, ours is based on the Lagrangian Decomposition of the nonlinear activation's convex relaxation. Thanks to the properties of Lagrangian Decomposition~\citep{Guignard1987}, we can show that our dual problem provides better bounds when evaluated on the same dual
variables. The relationship between the two duals is studied in detail in section~\ref{sec:dual_comp}.
Moreover, in terms of the followed optimization strategy, in addition to using a supergradient method like \citet{Dvijotham2018}, we present a proximal method, for which we can derive optimal step sizes. We show that these modifications enable us to compute tighter bounds using the same amount of compute time.


\section{PRELIMINARIES}
Throughout this paper, we will use bold lower case letters (like $\zb$) to
represent vectors and upper case letters (like $W$) to represent matrices.
Brackets are used to indicate the $i$-th coordinate of a vector ($\zb[i]$), and integer ranges (e.g., $[1,n-1]$). We
will study the computation of the lower bound problem based on a feedforward
neural network, with element-wise activation function $\sig{\cdot}$. The network
inputs are restricted to a convex domain $\mathcal{C}$, over which we assume
that we can easily optimise linear functions. This is the same assumption that
was made by \citet{Dvijotham2018}. The computation for an upper bound is
analogous. Formally, we wish to solve the following problem:
\begin{subequations}
	\begin{alignat}{2}
	\smash{\min_{\zb, \zbhat}}\quad & \zbhat_{n} \\
	\text{s.t. }\quad& \zb_0 \in \mathcal{C},\label{eq:verif-inibounds}\span\\
	& \zbhat_{k+1} = W_{k+1} \zb_{k} + \mathbf{b}_{k+1} \quad && k \in \left[1,n-1\right],\label{eq:verif-linconstr}\\
	& \zb_{k} = \sig{\zbhat_k} \quad &&k \in \left[1,n-1\right].\label{eq:verif-relu}
	\end{alignat}
	\label{eq:noncvx_pb}%
\end{subequations}
The output of the $k$-th layer of the network before and after the application of the activation function are denoted by $\zbhat_k$ and $\zb_k$ respectively.
Constraints \eqref{eq:verif-linconstr} implements the linear layers
(fully connected or convolutional) while constraints \eqref{eq:verif-relu}
implement the non-linear activation function.
Constraints~\eqref{eq:verif-inibounds} define the region over which the bounds are being computed. While our method can be extended to more complex networks (such as
ResNets), we focus on problem~\eqref{eq:noncvx_pb} for the sake of clarity.

The difficulty of problem \eqref{eq:noncvx_pb} comes from the
non-linearity~\eqref{eq:verif-relu}. \citet{Dvijotham2018} tackle it by relaxing \eqref{eq:verif-linconstr} and \eqref{eq:verif-inibounds} via Lagrangian multipliers, yielding the following dual:
\begin{equation}
\begin{split}
\max_{\mub,\lambdab}\enskip & d(\mub,\lambdab),  \quad \text{where: } \\
\hspace{-5pt}d(\mub, \lambdab) = & \min_{\zb, \zbhat} \quad W_n\zb_{n-1} + \mbf{b}_n \\
\qquad + & \sum_{k=1}^{n-1} \mub_k^T \left( \zbhat_k - W_k \zb_{k-1} - \mbf{b}_k \right) \\
\qquad + &\sum_{k=1}^{n-1} \lambdab_k^T \left( \zb_k - \sigma(\zbhat_k) \right)\\
\text{s.t. }\quad& \lb_k \leq \zbhat_k \leq \ub_k \qquad\qquad  k \in \left[1,n-1\right],\\
& \sigma(\lb_k) \leq \zb_k \leq \sigma(\ub_k) \qquad  k \in \left[1,n-1\right],\\
& \zb_0 \in \mathcal{C}.
\end{split}
\label{eq:dj_prob-main}
\end{equation}

If $\sigma$ is a ReLU, this relaxation is equivalent~\citep{Dvijotham2018} to the PLANET relaxation~\citep{Ehlers2017}.
The dual requires upper ($\ub_k$) and lower bounds
($\lb_k$) on the value that $\zbhat_k$ can take, for \mbox{$k \in \left[
	0..n-1\right]$}.  We call these \emph{intermediate bounds}: we detail how to compute them in appendix~\ref{sec:intermediate}.
The dual algorithm by \citet{Dvijotham2018} solves \eqref{eq:dj_prob-main} via supergradient ascent.



\section{LAGRANGIAN DECOMPOSITION}
We will now describe a novel approach to solve problem ~\eqref{eq:noncvx_pb}, and relate it to the dual algorithm by ~\citet{Dvijotham2018}.
We will focus on
computing bounds for an output of the last layer of the neural network.

\subsection{PROBLEM DERIVATION}
Our approach is based on Lagrangian decomposition, also known as variable splitting~\citep{Guignard1987}. Due to
the compositional structure of neural networks, most constraints involve only a limited number of variables. As a result, we can split the
problem into meaningful, easy to solve subproblems. We then impose
constraints that the solutions of the subproblems should agree.

We start from the original non-convex primal problem \eqref{eq:noncvx_pb}, and substitute the nonlinear activation equality \eqref{eq:verif-relu} with a constraint corresponding to its convex hull.
This leads to the following convex program:
\begin{subequations}
	\begin{alignat}{2}
	\hspace{-5pt}\smash{\min_{\zb, \zbhat}}\enskip & \zbhat_n \\
	\hspace{-5pt}\text{s.t. }\enskip& \zb_0 \in \mathcal{C}, \label{eq:planet-inibounds}\span\\
	& \zbhat_{k+1} = W_{k+1} \zb_{k} + \mathbf{b}_{k+1}\enskip &&  k \in \left[1,n-1\right],\label{eq:planet-linconstr}\\
	& \texttt{cvx\_hull}_\sigma\left( \zbhat_k, \zb_k, \lb_k, \ub_k\right) \enskip &&k \in \left[1,n-1\right].\label{eq:planet-acthull}
	\end{alignat}
	\label{eq:planet-relax}%
\end{subequations}
In the following, we will use ReLU activation functions as an example, and employ the PLANET relaxation as its convex hull, which is the tightest linearly sized relaxation published thus far. By linearly sized, we mean that the numbers of
variables and constraints to describe the relaxation only grow
linearly with the number of units in the network.
We stress that the derivation can be extended to other
non-linearities. For example, appendix \ref{sec:sigmoid_act} describes the case
of sigmoid activation function.
For ReLUs, this convex hull takes the following form~\citep{Ehlers2017}:
\begin{subnumcases}
{\label{eq:planet-relu-hull}\hspace{-30pt}\texttt{cvx\_hull}_\sigma\equiv}
\text{if } \lb_k[i] \leq  0; \ub_k[i] \geq 0: \nonumber \\
\quad \zb_{k}[i] \geq 0, \quad \zb_k[i] \geq \zbhat_k[i],  \label{eq:planet-reluamb}\\
\quad \zb_{k}[i] \leq \frac{\mbf{u_k}[i](\zbhat_k[i] - \mbf{l}_k[i])}{\mbf{u_k}[i]-\mbf{l_k}[i]}. \nonumber\\
\text{if } \ub_k[i] \leq 0: \nonumber \\
\quad \zb_k[i] = 0. \label{eq:planet-relublock}\\
\text{if } \lb_k[i] \geq 0: \nonumber \\
\quad \zb_k[i] = \zbhat_k[i]. \label{eq:planet-relupass}
\end{subnumcases}
Constraints~\eqref{eq:planet-reluamb}, \eqref{eq:planet-relublock}, and
\eqref{eq:planet-relupass} corresponds to the relaxation of the ReLU
\eqref{eq:verif-relu} in different cases (respectively: ambiguous state, always blocking or always passing).

To obtain a decomposition, we divide the constraints into subsets. Each subset
will correspond to a pair of an activation layer, and the linear layer coming
after it. The only exception is the first linear layer which is combined with
the restriction of the input domain to $\mathcal{C}$. This choice is motivated
by the fact that for piecewise-linear activation functions, we will be able to
easily perform linear optimisation over the resulting subdomains. For a
different activation function, it might be required to have a different
decomposition. Formally, we introduce the following notation for subsets of constraints:
\begin{equation}
	\hspace{-.2\linewidth}
	\mathcal{P}_0(\zb_{0}, \zbhat_{1}) \equiv \begin{cases} \zb_{0} \in \mathcal{C}\\
	\zbhat_{1} = W_{1} \zb_{0} + \mathbf{b}_{1},
	\end{cases}
	\hspace{-.2\linewidth}
	\label{eq:pb1}
\end{equation}
\begin{equation}
	\hspace{-.3\linewidth}
	\mathcal{P}_k(\zbhat_{k}, \zbhat_{k+1}) \equiv
	\begin{cases}
	\lb_{k} \leq \zbhat_{k} \leq \ub_{k},\span\\
	\texttt{cvx\_hull}_\sigma\left( \zbhat_k, \zb_k, \lb_k, \ub_k\right),\\
	\zbhat_{k+1} = W_{k+1} \zb_{k} + \mathbf{b}_{k}.
	\end{cases}
	\label{eq:pbk}
\end{equation}

The set $\mathcal{P}_k$ is defined without an explicit dependency on $\zb_k$
because $\zb_k$ only appears internally to the subset of constraints and is not
involved in the objective function. Using this grouping of the constraints, we
can concisely write problem \eqref{eq:planet-relax} as:
\begin{equation}
\begin{split}
\smash{\min_{\zb, \zbhat}}&\quad \zbhat_n \qquad \text{s.t. }\quad  \mathcal{P}_0(\zb_0, \zbhat_{1}), \\
&\mathcal{P}_k(\zbhat_{k}, \zbhat_{k+1}) \qquad k \in \left[1,n-1\right].
\end{split}
\label{eq:nondec_pb}
\end{equation}
To obtain a Lagrangian decomposition, we duplicate the variables so that each
subset of constraints has its own copy of the variables it is involved in. Formally, we
rewrite problem~\eqref{eq:nondec_pb} as follows:
\begin{subequations}
	\begin{alignat}{2}
	\smash{\min_{\zb, \zbhat}}&\quad \zbhat_{A,n} \\[3pt]
	\text{s.t. } & \quad \mathcal{P}_0(\zb_0, \zbhat_{A,1}), \\
	&\quad\mathcal{P}_k(\zbhat_{B, k}, \zbhat_{A, k+1}) &&\qquad k \in \left[1,n-1\right],\\
	&\quad \zbhat_{A, k} = \zbhat_{B, k} &&\qquad k \in \left[1,n-1\right]. \label{eq:matcheq}%
	\end{alignat}
	\label{eq:primal_lag_problem}%
\end{subequations}
The additional equality constraints \eqref{eq:matcheq} impose agreements between
the various copies of variables. We introduce the dual variables $\lambdabhat$
and derive the Lagrangian dual:
\vspace{-10pt}
\begin{equation}
\begin{split}
\max_{\lambdabhat}\enskip & q(\lambdabhat),  \quad \text{where: } \\[-4pt]
q\left( \lambdabhat \right) &= \min_{\zb, \zbhat}\ \zbhat_{A, n}
+ \sum_{k=1}^{n-1} \lambdabhat_k^T \left( \zbhat_{B, k} - \zbhat_{A, k} \right)\\[5pt]
\text{s.t. }\quad & \mathcal{P}_0(\zb_0, \zbhat_{A,1}), \\
 &\mathcal{P}_k(\zbhat_{B, k}, \zbhat_{A, k+1}) \qquad k \in \left[1,n-1\right].
\end{split}
\label{eq:nonprox_problem}
\end{equation}
Any value of $\lambdabhat$ provides a valid lower bound by
virtue of weak duality. While we will maximise over the choice of dual variables
in order to obtain as tight a bound as possible, we will be able to
interrupt the optimisation process at any point and obtain a valid bound by
evaluating $q$.

We stress that, at convergence, problems \eqref{eq:nonprox_problem} and \eqref{eq:dj_prob-main} will yield the same bounds, as they are both reformulations of the same convex problem.
This does not imply that the two derivations yield solvers with the same efficiency. In fact, we will next prove that, for ReLU-based networks, problem \eqref{eq:nonprox_problem} yields bounds at least as tight as a corresponding dual solution from problem \eqref{eq:dj_prob-main}.

\subsection{DUALS COMPARISON} \label{sec:dual_comp}
We now compare our dual problem \eqref{eq:nonprox_problem} with problem \eqref{eq:dj_prob-main} by \citet{Dvijotham2018}. From the high level perspective, our decomposition considers larger subsets of constraints and hence results in a smaller number of variables to optimize. We formally prove that, for ReLU-based neural networks, our
formulation dominates theirs, producing tighter bounds based on the same dual
variables.
\begin{theorem} 
	Let us assume $\sigma(\zbhat_k) = \max(\mathbf{0}, \zbhat_k)$. For any solution $\mub, \lambdab$ of dual \eqref{eq:dj_prob-main} by \citet{Dvijotham2018} yielding bound $d(\mub, \lambdab)$, it holds that $q(\mub) \geq d(\mub, \lambdab)$.
	\label{theorem}
\end{theorem}
\begin{proof}
	See appendix~\ref{sec:dual_comp_proof}.
\end{proof}
Our dual is also related to the one that \citet{Wong2018} operates in. We show in
appendix~\ref{sec:WK_init} how our problem can be initialised to a set of dual
variables so as to generate a bound matching the one provided by their
algorithm. We use this approach as our initialisation for incomplete verification ($\S$ \ref{sec:experiments-incomplete}).

\section{SOLVERS FOR LAGRANGIAN DECOMPOSITION}
Now that we motivated the use of dual \eqref{eq:nonprox_problem} over problem \eqref{eq:dj_prob-main}, it remains to show how to solve it efficiently in practice. We present two methods: one based on supergradient ascent, the other on proximal maximisation.
A summary, including pseudo-code, can be found in appendix~\ref{sec:algo_summary}.

\subsection{SUPERGRADIENT METHOD}
\label{subsec:subgradient_method}
As \citet{Dvijotham2018} use supergradient methods on their dual, we start by applying it on problem \eqref{eq:nonprox_problem} as well.

At a given point $\lambdabhat$, obtaining the supergradient requires us to know
the values of $\zbhat_{A}$ and $\zbhat_{B}$ for which the inner minimisation is
achieved. Based on the identified values of $\zbhat_{A}^*$ and $\zbhat_{B}^*$, we can then compute
the supergradient \mbox{$\nabla_{\lambdabhat} q= \zbhat_{B}^* - \zbhat_{A}^*$}.
The updates to $\lambdabhat$ are then the usual supergradient ascent updates:
\begin{equation}
\lambdabhat^{t+1} = \lambdabhat^{t} + \alpha^t \nabla_{\lambdabhat} q(\lambdabhat^t),
\label{eq:subgradient_update}
\end{equation}
where $\alpha^t$ corresponds to a step size schedule that needs to be provided.
It is also possible to use any variants of gradient descent, such as
Adam~\citep{Kingma2015}.

It remains to show how to perform the inner minimization over the primal variables.
By design, each of the
variables is only involved in one subset of constraints. As a result, the
computation completely decomposes over the
subproblems, each corresponding to one of the subset of constraints. We
therefore simply need to optimise linear functions over one subset of
constraints at a time. 

\subsubsection{Inner minimization: $\mathcal{P}_0$ subproblems} \label{sec:innerminp0}
To minimize over $\zb_{0}, \zbhat_{A, 1}$, the variables constrained by $\mathcal{P}_0$, we need to solve:
\begin{equation}
\begin{split}
[\zb_{0}^*, \zbhat_{A, 1}^*] &= \argmin_{\zb_{0}, \zbhat_{A, 1}}\ -\lambdabhat_1^T \zbhat_{A, 1} \\
\text{s.t } \quad &\zb_0 \in \mathcal{C}, \qquad \zbhat_{A, 1} = W_{1} \zb_0 + \mathbf{b}_{0}.
\end{split}
\label{eq:typeA_cgd}
\end{equation}
Rewriting the problem as a linear function of $\zb_{0}$ only, problem
\eqref{eq:typeA_cgd} is simply equivalent to minimising $-\lambdabhat_1^T W_1 \zb_0$
over $\mathcal{C}$. We assumed that the optimisation of a linear function
over $\mathcal{C}$ was efficient. We now give examples of $\mathcal{C}$ where problem
\eqref{eq:typeA_cgd} can be solved efficiently.

\paragraph{Bounded Input Domain}
If $\mathcal{C}$ is defined by a set of lower bounds $\lb_0$ and upper bounds
$\ub_0$ (as in the case of $\ell_{\infty}$ adversarial examples),
optimisation will simply amount to choosing either the lower or upper bound
depending on the sign of the linear function. Let us denote the indicator function for condition $c$ by $\mathbb{1}_c$. The optimal solution is:
\begin{equation}
\begin{split}
\xb_{0} &=\  \mathbb{1}_{\lambdabhat_1^T W_1 < 0} \cdot \lb_{0} +\mathbb{1}_{\lambdabhat_1^T W_1 \geq 0} \cdot \ub_0, \\
\xbhat_{A, 1} &=\ W_{1} \xb_{0} + \mbf{b}_{1}.
\end{split}
\label{eq:box_bounds_sol}
\end{equation}

\paragraph{$\ell_2$ Balls} If $\mathcal{C}$ is defined by an
$\ell_2$ ball of radius $\epsilon$ around a point $\mbf{\bar{x}}$
($\|\xb_0 - \mbf{\bar{x}}\|_2 \leq \epsilon$), optimisation amounts to choosing
the point on the boundary of the ball such that the vector from the center to it
is opposed to the cost function. Formally, the optimal solution is given by:
\begin{equation}
\begin{split}
\xbhat_{A, 1} &=\  W_{1} \xb_{0} + \mbf{b}_{1}, \\
\xb_0 &=\ \mbf{\bar{x}} + (\nicefrac{\sqrt{\epsilon}}{\|\lambdabhat_1\|_2}) \lambdabhat_1.
\end{split}
\end{equation}

\subsubsection{Inner minimization: $\mathcal{P}_k$ subproblems} \label{sec:innerminpk}
For the variables constrained by subproblem $\mathcal{P}_k$ ($\zbhat_{B, k}, \zbhat_{A, k+1}$), we need to solve:
\begin{equation}
\begin{split}
\hspace{-7pt}[\zbhat_{B, k}^*&, \zbhat_{A, k+1}^*] = \smash{\argmin_{\mathclap{\zbhat_{B, k}, \zbhat_{A, k+1}}}} \quad
\lambdabhat_k^T \zbhat_{B, k} - \lambdabhat_{k+1}^T \zbhat_{A, k+1} \\[8pt]
\text{s.t } & \quad \lb_{k} \leq \zbhat_{B, k} \leq \ub_{k},\\
& \quad \texttt{cvx\_hull}_\sigma\left( \zbhat_{B,k}, \zb_k, \lb_k, \ub_k\right),\\
& \quad \zbhat_{A, k+1} = W_{k+1} \zb_{k} + \mathbf{b}_{k+1}.
\end{split} \label{eq:pk_sub}
\end{equation}
\begin{figure}[t!]
	\vskip -12pt
	\centering
	\noindent\resizebox{.28\textwidth}{!}{
		\begin{tikzpicture}
  \tikzset{dummy/.style= {inner sep=0, outer sep=0}}
  \tikzset{cross/.style={circle, draw,
      minimum size=5*(#1-\pgflinewidth),
      inner sep=0pt, outer sep=0pt,
      ultra thick, color=red}}

  \draw[-, ultra thick](-1, 0) to (0, 0) to (1, 1);

  \draw[dashed](-1, -0.5) to (-1, 1.5);
  \draw[dashed](1, -0.5) to (1, 1.5);

  \draw[fill=green](-1, 0) -- (0,0) -- (1, 1);

  \node[dummy](lb-lab) at (-1.3, -0.3) {$l_{k[j]}$};
  \node[dummy](ub-lab) at (1.35, -0.3) {$u_{k[j]}$};

  \node[cross=2pt] at (-1, 0) {};
  \node[cross=2pt] at (1, 1) {};
  \node[cross=2pt] at (0, 0) {};

  \draw[-latex](-1.5,0) to (2, 0);
  \node[dummy](x-label) at (2.3, 0) {$\zbhat_{k[j]}$};
  \draw[-latex](0,-0.5) to (0, 1.5);
  \node[dummy](x-label) at (0, 1.8) {$\zb_{k[j]}$};

\end{tikzpicture}
	}
	\caption{Feasible domain of the convex hull for an ambiguous ReLU. Red
		circles indicate the vertices of the feasible region.}
	\label{fig:relucvx}
\end{figure}
\hspace{-3pt}In the case where the activation function $\sigma$ is the ReLU and
$\texttt{cvx\_hull}$ is given by equation~\eqref{eq:planet-relu-hull}, we can
find a closed form solution. Using the last equality of the problem, we can
rewrite the objective function as \mbox{$\lambdabhat_k^T \zbhat_{B, k} - \lambdabhat_{k+1}^T W_{k+1} \zb_{k}$}. If the ReLU is ambiguous, the shape of the
convex hull is represented in Figure~\ref{fig:relucvx}. If the sign of
$\lambdabhat_{k+1}^T W_{k+1}$ is negative, optimizing subproblem~\eqref{eq:pk_sub} corresponds
to having $\zb_{k}$ at its lower bound \mbox{$\max\left(\zbhat_{B,k},
	0\right)$}. If on the contrary the sign is positive, $\zb_{k}$ must be
equal to its upper bound \mbox{$\frac{\ub_k}{\ub_k-\lb_k}\left( \zbhat_{B, k} -
	\lb_k \right)$}. We can therefore rewrite the subproblem~\eqref{eq:pk_sub}
in the case of ambiguous ReLUs as:
\begin{equation}
\begin{split}
\hspace{-10pt}\argmin_{\zbhat_{B,k} \in \left[ \lb_k, \ub_k \right]} \enskip & \left(\lambdabhat_{k}^T - \left[ \lambdabhat_{k+1}^TW_{k+1} \right]_{+} \frac{\ub_k}{\ub_k - \lb_k}\right) \zbhat_{B, k} \\
&- \left[ \lambdabhat_{k+1}^TW_{k+1} \right]_{-} \max\left( \zbhat_{B, k}, 0\right),
\end{split}
\label{eq:ambrelu_linopt}
\end{equation}
where $[A]_{-}$ corresponds to the negative values of $A$ ($[A]_{-} = \min\left(
A, 0 \right)$) and $[A]_{+}$ corresponds to the positive value of A ($[A]_{+}
= \max\left( A, 0\right)$). The objective function and the constraints all
decompose completely over the coordinates, so all problems can be solved
independently. For each dimension, the problem is a convex, piecewise linear
function, which means that the optimal point will be a vertex. The possible
vertices are \mbox{$\left( \lb_k[i], 0 \right)$}, \mbox{$\left( 0, 0 \right)$},
and \mbox{$\left( \ub_k[i], \ub_k[i] \right)$}. In order to find the minimum, we can therefore 
evaluate the objective function at these three points and keep the one with the smallest value.

If for a ReLU we either have $\lb_k[i] \geq 0$ or $\ub_k[i] \leq 0$, the $\texttt{cvx\_hull}$
constraint is a simple linear equality constraint. For those coordinates, the
problem is analogous to the one solved by equation~\eqref{eq:box_bounds_sol},
with the linear function being minimised over the $\zbhat_{B,k}$ box bounds
being $\lambdabhat_k^T \zbhat_{B,k}$ in the case of blocking ReLUs or \mbox{$\left(
	\lambdabhat_{k}^T - \lambdabhat_{k+1}^TW_{k+1}\right) \zbhat_{B, k}$} in the case of
passing ReLUs.

We have described how to solve problem~\eqref{eq:pk_sub} by simply evaluating
linear functions at given points. The matrix operations involved correspond to standard operations of the neural network's layers. Computing
\mbox{$\smash{\zbhat_{A,k+1} = W_{k+1} \zb_{k} + \mathbf{b}_{k+1}}$} is exactly
the forward pass of the network and computing $\smash{\lambdabhat_{k+1}^T
	W_{k+1}}$ is analogous to the backpropagation of gradients. We can therefore
take advantage of existing deep learning frameworks to gain access to efficient
implementations. When dealing with convolutional layers, we can employ
specialised implementations rather than building the equivalent $W_k$ matrix, which would contain a lot of redundancy.

We described here the solving process in the context of ReLU activation
functions, but this can be generalised to non piecewise linear activation
function. For example, appendix \ref{sec:sigmoid_act} describes the solution for
sigmoid activations.

\subsection{PROXIMAL METHOD} \label{sec:prox}

We now detail the use of proximal methods on problem \eqref{eq:nonprox_problem} as an alternative to supergradient ascent.

\subsubsection{Augmented Lagrangian}
Applying proximal maximization to the dual function $q$ results in the Augmented
Lagrangian Method, also known as the method of multipliers. The derivation of
the update steps detailed by~\citet{Bertsekas1989}. For our problem,
this will correspond to alternating between updates to the dual variables
$\lambdabhat$ and updates to the primal variables $\zbhat$, which are given by
the following equations (superscript $t$ indicates the value at the
$t$-th iteration):
\vspace{-7pt}
\begin{gather}
\lambdabhat_k^{t+1} = \lambdabhat_k^{t} + \frac{\zbhat^{t}_{B,k} - \zbhat^{t}_{A,k}}{\eta_k} \label{eq:lambda_updates},\\[8pt]
\begin{split}
\left[ \zb^{t}, \zbhat^{t} \right] \ &= \smash{\argmin_{\zb, \zbhat}}\, \mathcal{L}\left( \zbhat, \lambdabhat^t\right) \\
 &:=\smash{\argmin_{\zb, \zbhat}} \Bigg[ \enskip \zbhat_{A, n} + \sum_{k=1..n-1} \lambdabhat_k^T \left( \zbhat_{B, k} - \zbhat_{A, k} \right) \\
& \quad + \sum_{k=1..n-1} \frac{1}{2 \eta_k}  \| \zbhat_{B, k} - \zbhat_{A, k}\|^2 \Bigg]\\
\text{s.t. } \quad & \mathcal{P}_0(\zb_0, \zbhat_{A,1}), \\
& \mathcal{P}_k(\zbhat_{B, k}, \zbhat_{A, k+1})\ \quad k \in \left[1,n-1\right].
\end{split}
\label{eq:z_updates}
\end{gather}
The term $\auglag$ is the Augmented Lagrangian of problem \eqref{eq:primal_lag_problem}.
The additional quadratic term in \eqref{eq:z_updates}, compared to the objective of $q(\lambdabhat)$,
arises from the proximal terms on $\lambdabhat$. It has the advantage of
making the problem strongly convex, and hence easier to optimize. Later on,
will show  that this allows us to derive optimal
step-sizes in closed form. The weight $\eta_k$ is a hyperparameter of the algorithm. A high
value will make the problem more strongly convex and therefore quicker to solve,
but it will also limit the ability of the algorithm to perform large updates.

While obtaining the new values of $\lambdabhat$ is trivial using
equation~\eqref{eq:lambda_updates}, problem \eqref{eq:z_updates} does not have a
closed-form solution. We show how to solve it efficiently
nonetheless.

\subsubsection{Frank-Wolfe Algorithm}
\label{subsec:FW}
Problem~\eqref{eq:z_updates} can be optimised using the conditional gradient
method, also known as the Frank-Wolfe algorithm~\citep{Frank1956}. The advantage
it provides is that there is no need to perform a projection step to remain in
the feasible domain. Indeed, the different iterates remain in the feasible
domain by construction as convex combination of points in the feasible domain.
At each time step, we replace the objective by its linear approximation and
optimize this linear function over the feasible domain to get an update
direction, named conditional gradient. 
We then take a step in this direction. As the Augmented Lagrangian is smooth over the primal variables, there is no need to take the Frank-Wolfe step for all the network layers at once. We can in fact do it in a block-coordinate fashion, where a block is a network layer, with the goal of speeding up convergence; we refer the reader to appendix \ref{sec:prox-appendix} for further details.

Obtaining the conditional gradient requires minimising a linearization of $\auglag$ on the primal variables, restricted to the feasible domain. 
This computation corresponds exactly to the one we do to perform the inner minimisation of problem \eqref{eq:nonprox_problem} over $\zb$ and $\zbhat$ in order to compute the supergradient (cf. $\S$\ref{sec:innerminp0}, $\S$\ref{sec:innerminpk}). To make this equivalence clearer, we point out that the linear coefficients of the primal variables will maintain the same form (with the difference that the dual variables are represented as their closed-form update for the following iteration), as \mbox{$\nabla_{\zbhat_{B, k}} \mathcal{L}\left( \zbhat,\lambdabhat^t\right) = \lambdabhat_{k}^{t+1}$} and \mbox{$\nabla_{\zbhat_{A, k+1}}\mathcal{L}\left( \zbhat,\lambdabhat^t\right) = -\lambdabhat_{k}^{t+1}$}.
The equivalence of conditional gradient and supergradient is not particular to our problem. A more general description can be found in the work of \citet{Bach2015}.

The use of a proximal method allows us to employ an optimal step size, whose calculation is detailed in appendix \ref{sec:prox-appendix}. This would not be possible with supergradient methods, as we would have no guarantee of improvements.
We therefore have no need to choose the step-size. In practice, we still have to
choose the strength of the proximal term $\eta_k$. 
Finally, inspired by previous work on accelerating proximal methods \citep{Lin2017,Salzo12}, we also apply momentum on the dual updates to accelerate convergence; for details see appendix \ref{sec:prox-appendix}.

\section{EXPERIMENTS}
\subsection{IMPLEMENTATION DETAILS} \label{sec:implementation}
One of the benefits of our algorithms (and of the supergradient baseline by \citet{Dvijotham2018}) is that
they are easily parallelizable. As explained in
$\S$~\ref{sec:innerminp0} and $\S$~\ref{sec:innerminpk}, the crucial part of both our solvers works by applying linear algebra
operations that are equivalent to the ones employed during the forward and backward
passes of neural networks. 
To compute the upper and lower bounds for all
the neurons of a layer, there is no dependency between the different problems,
so we are free to solve them all simultaneously in a batched mode. 
In complete verification ($\S$\ref{sec:experiments-complete}), where bounds relative to multiple domains (defined by the set of input and intermediate bounds for the given network activations) we can further batch over the domains, providing a further stream of parallelism.
In practice, we take the building blocks
provided by deep learning frameworks, and apply them to batches of solutions
(one per problem), in the same way that normal training applies them to a batch
of samples. This makes it possible for us to leverage the engineering efforts
made to enable fast training and evaluation of neural networks, and easily take
advantage of GPU accelerations. The implementation used in our experiments is
based on Pytorch~\citep{Paszke2017}. 

\begin{figure*}[b!]
	\vspace{-2pt}
	\centering
		\caption{\label{fig:madry8} Distribution of runtime
		and gap to optimality on a network adversarially trained with the method by ~\citet{Madry2018}, on CIFAR-10 images. In both
		cases, lower is better. The width at a given value represents the
		proportion of problems for which this is the result. Gurobi
		always return the optimal solution so doesn't appear on the optimality
		gap, but is always the highest runtime. \vspace{-3pt}}
	\includegraphics[width=1\textwidth]{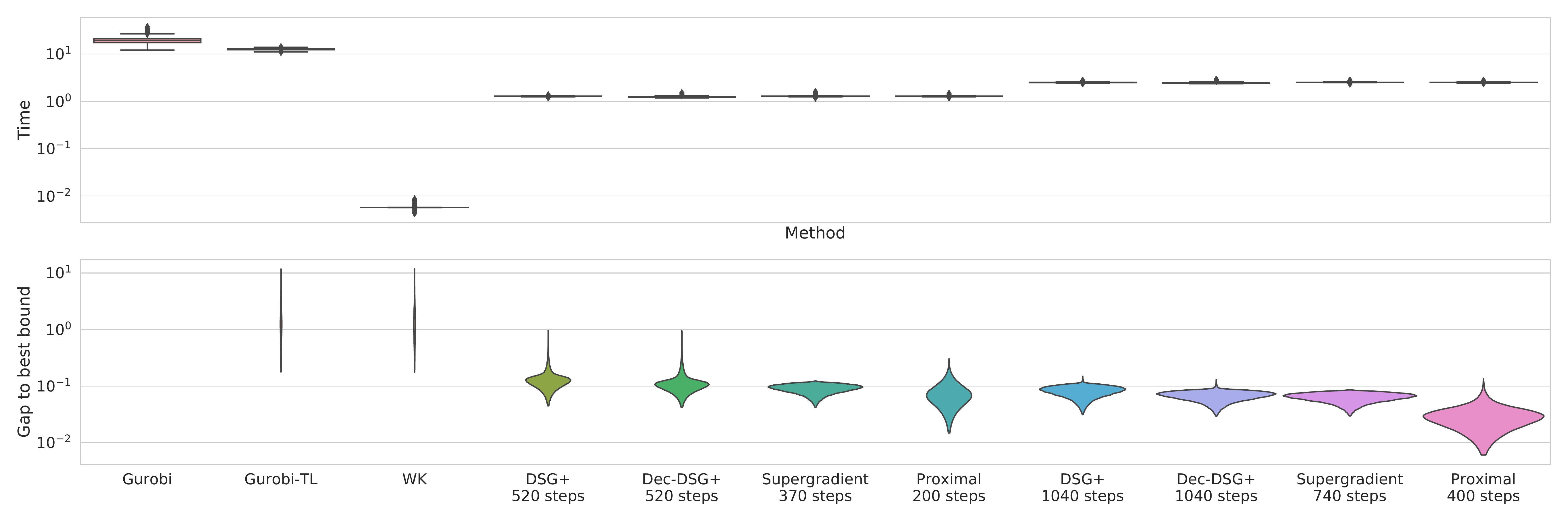}
\end{figure*}

\subsection{EXPERIMENTAL SETTING} \label{sec:experiments-setting}
In order to evaluate the performance of the proposed methods, we first perform a
comparison in the quality and speed of bounds obtained by different methods in the context of incomplete verification ($\S$\ref{sec:experiments-incomplete}). We then assess the effect of the various method-specific speed/accuracy trade-offs within a complete verification procedure ($\S$\ref{sec:experiments-complete}).

For both sets of experiments, we report results using networks trained with
different methodologies on the CIFAR-10 dataset. 
We compare the following algorithms:
\begin{itemize}[nosep,leftmargin=1em,labelwidth=*,align=left]
	\item {\bf WK} uses the method of \citet{Wong2018}.
	This is equivalent to a specific choice of $\lambdabhat$.
	\item {\bf DSG+} corresponds to supergradient ascent on dual \eqref{eq:dj_prob-main}, the method by \citet{Dvijotham2018}. 
	We use the Adam~\citep{Kingma2015}
	updates rules and decrease the step size linearly, similarly to the experiments of \citet{Dvijotham2018}. We
	experimented with other step size schedules, like constant step size or
	$\frac{1}{t}$ schedules, which all performed worse. 
	\item  {\bf Dec-DSG+} is a direct application of Theorem~\ref{theorem}: it obtains a dual solution for problem \eqref{eq:dj_prob-main} via DSG+ and sets $\lambdabhat=\mub$ in problem~\eqref{eq:nonprox_problem} to obtain the final bounds.
	\item {\bf Supergradient} is the first of the two solvers presented ($\S$\ref{subsec:subgradient_method}).
	It corresponds to supergradient ascent over
	problem~\eqref{eq:nonprox_problem}. We use Adam updates
	as in DSG+.
	\item {\bf Proximal} is the solver presented in $\S$\ref{sec:prox}, performing proximal maximisation on problem~\eqref{eq:nonprox_problem}.
	We limit the total number of inner iterations for each problem to $2$ (a single iteration is a full pass over the network layers). 
	\item {\bf Gurobi} is our gold standard method, employing the commercial black
	box solver Gurobi. All the problems are solved to optimality, providing us
	with the best result that our solver could achieve in terms of accuracy.
	\item {\bf Gurobi-TL} is the time-limited version of the above, which stops at the first dual simplex iteration for which the total elapsed time exceeded that required by $400$ iterations of the proximal method.
\end{itemize}
\iftoggle{longversion}{ We chose to not include computation based on {\bf Naive
		bounds} in our analysis. This is the simplest method imaginable, based on
	interval propagation. It has been described as the Box
	domain~\citep{Mirman2018} or as Interval Bound Propagation~\citep{Gowal2018b},
	and corresponds to setting all the dual variables to 0 in our dual
	formulation. While~\citep{Gowal2018b} showed that it could produce state of the
	art results when used as a training signal, our experiments showed it to
	produce bounds significantly worse than other methods for networks that were
	not trained with it. }{
	We omit {\bf Interval Bound Propagation}~\citep{Gowal2018b,Mirman2018} from the comparisons as on average it performed
	significantly worse than WK on the networks we used: experimental results are provided in appendix \ref{sec:experiments-incomplete-appendix}.}
For all of the methods above, as done in previous work for neural network verification \citep{Lu2020, Bunel2020}, we compute intermediate bounds (see appendix \ref{sec:intermediate}) by taking the layer-wise best bounds output by Interval Propagation and WK. This holds true for both the incomplete and the complete verification experiments.
Gurobi was run on $4$ CPUs for $\S$~\ref{sec:experiments-incomplete} and on $1$ for $\S$~\ref{sec:experiments-complete} in order to match the setting by \citet{Bunel2020}, whereas all the other methods were run on a single GPU. While it may seem to
lead to an unfair comparison, the flexibility and parallelism of the methods are
a big part of their advantages over off-the-shelf solvers.

\subsection{INCOMPLETE VERIFICATION} \label{sec:experiments-incomplete}

We investigate the effectiveness of the various methods for incomplete verification on images of the CIFAR-10 test set. For each image, we compute an upper bound on the robustness margin: the difference between the ground truth logit and all the other logits, under an allowed perturbation $\epsilon_{\text{verif}}$ in infinity norm of the inputs. If for any class the upper bound on the robustness margin is negative, then we are certain that the network is vulnerable against that adversarial perturbation. We measure the time to compute last layer bounds, and report the optimality gap compared to the best achieved solution.

As network architecture, we employ the small model used by
\citet{Wong2018} and whose structure is given in appendix \ref{sec:network_arch}. 
We train the network against perturbations of a size up to \mbox{$\epsilon_{\text{train}}=8/255$}, and test for adversarial vulnerability on $\epsilon_{\text{verif}} = 12/255$. This is done using adversarial training~\citep{Madry2018}, based on an
attacker using 50 steps of projected gradient descent to obtain the
samples. Additional experiments for a network trained using standard stochastic
gradient descent and cross entropy, with no robustness objective can be found in appendix \ref{sec:experiments-incomplete-appendix}.
For both supergradient methods (our Supergradient, and DSG+), we decrease the step size linearly from $\alpha=10^{-2}$ to $\alpha=10^{-4}$, while for Proximal, we employ momentum coefficient $\mu=0.3$ and, for all layers, linearly increase the weight of the proximal terms from $\eta=10^1$ to $\eta=5\times10^2$ (see appendix \ref{sec:prox-appendix}).

\begin{figure*}[t!]
	\centering
	\begin{subfigure}{.48\textwidth}
		\centering
		\includegraphics[width=\textwidth]{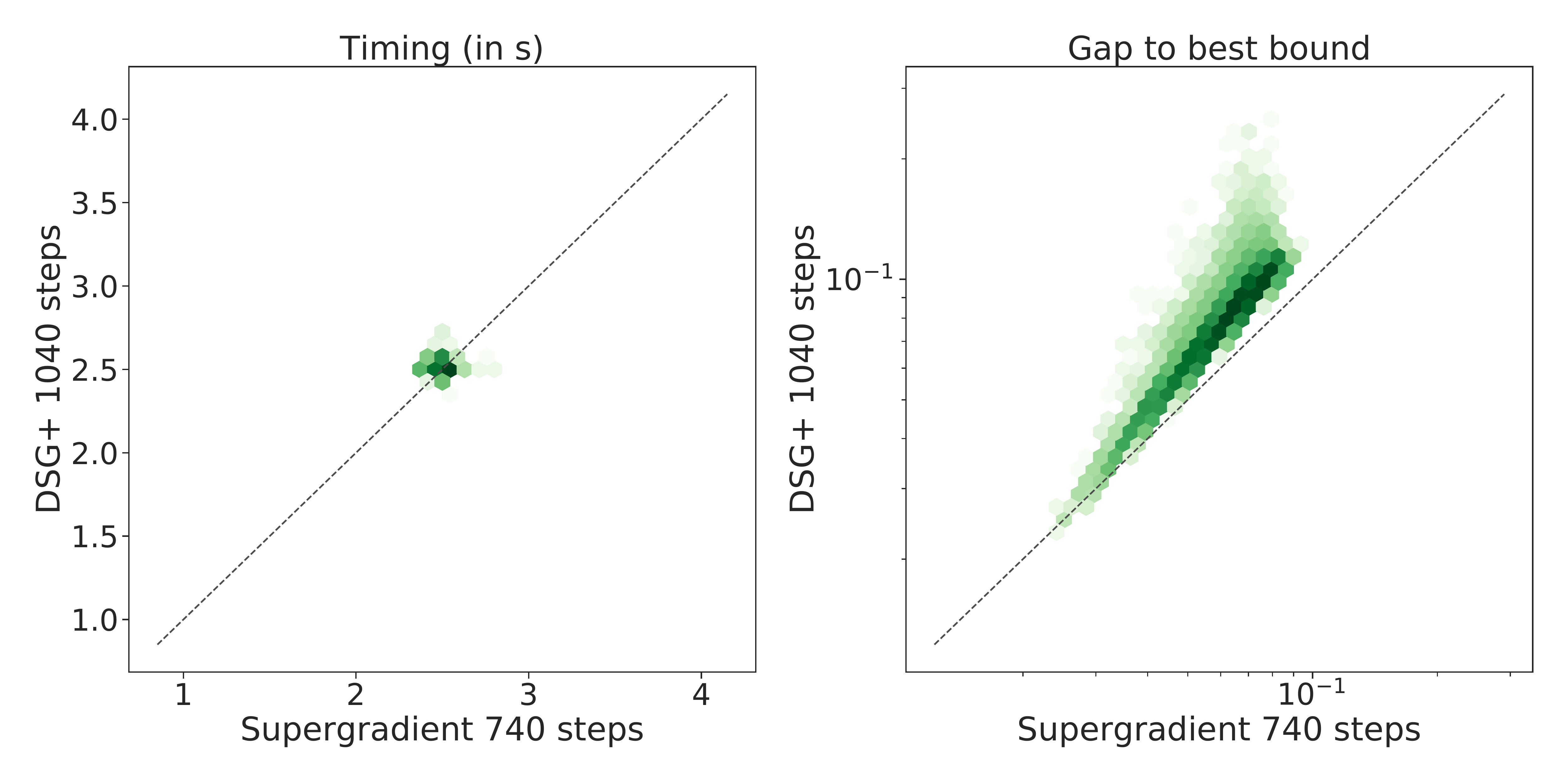}
	\end{subfigure}
	\begin{subfigure}{.48\textwidth}
		\centering
		\includegraphics[width=\textwidth]{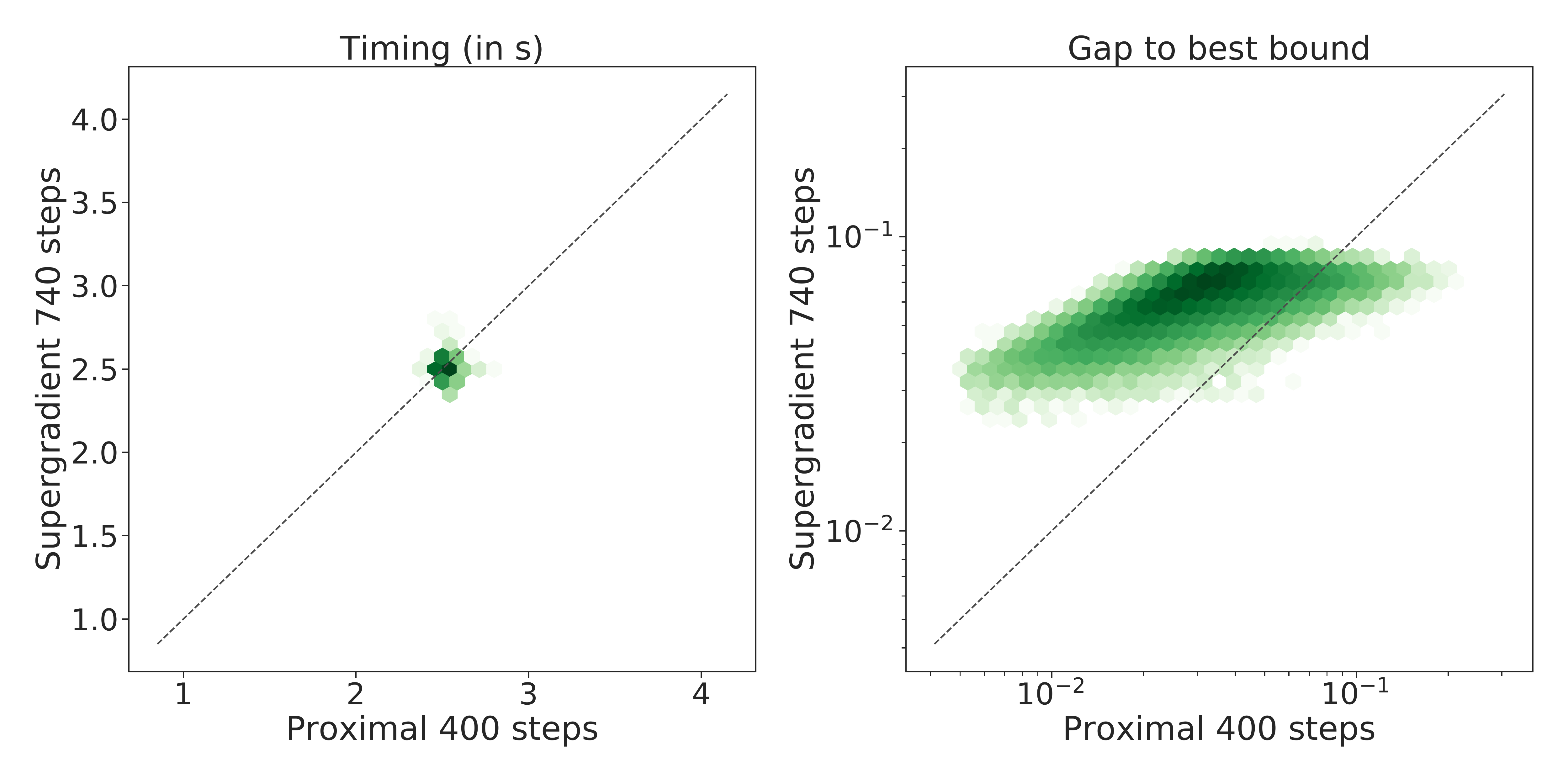}
	\end{subfigure}
	\caption{Pointwise comparison for a subset of the methods on the data presented in Figure \ref{fig:madry8}. Each datapoint corresponds to a CIFAR image, darker colour shades mean higher point density. The dotted line corresponds to the equality and in both graphs, lower is better along both axes.}
	\label{fig:madry8-pointwise}
	\vspace{-5pt}
\end{figure*}

\begin{figure*}[b!]
		\vspace{-15pt}
	\centering
	\begin{subfigure}{.32\textwidth}
		\centering
		\includegraphics[width=\textwidth]{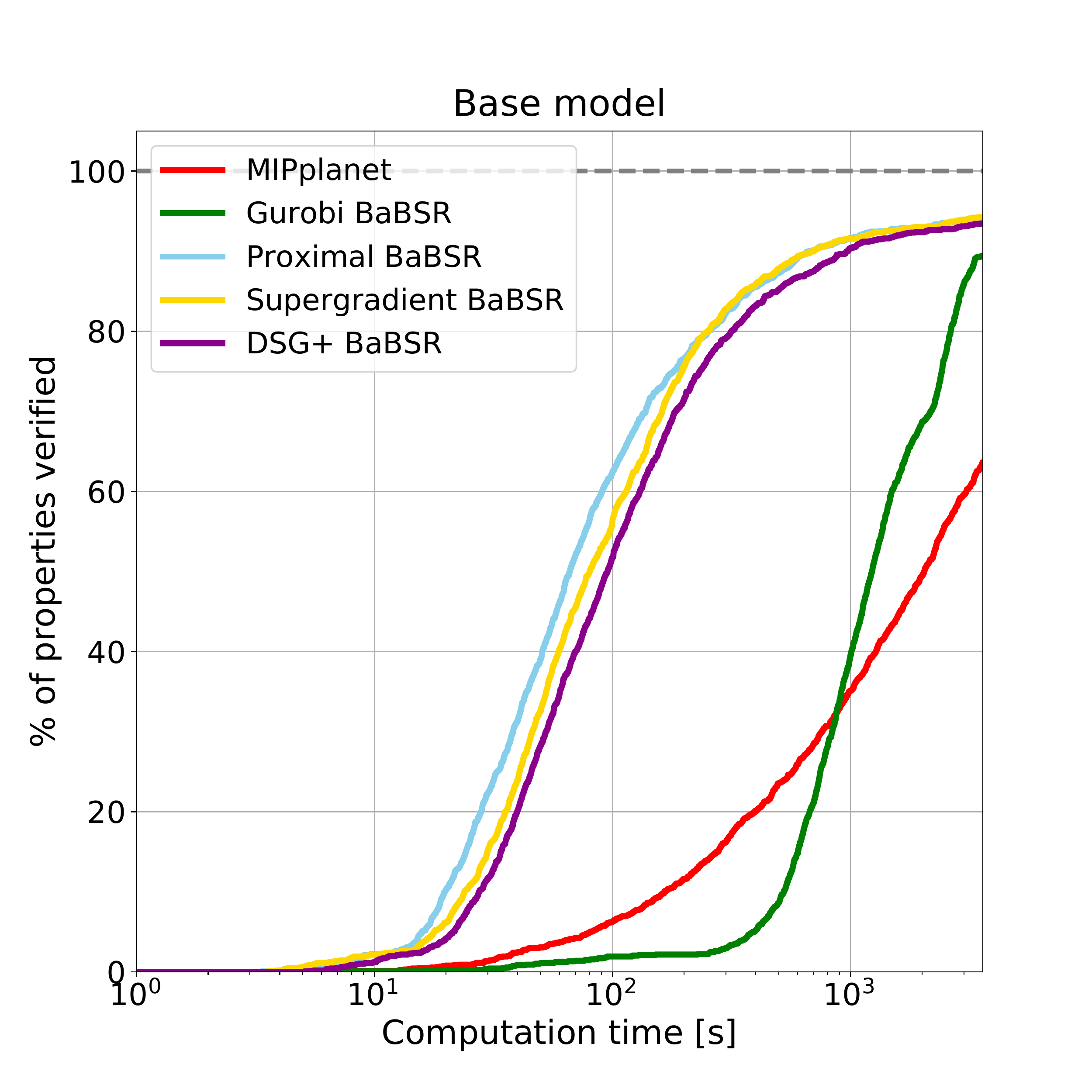}
	\end{subfigure}
	\begin{subfigure}{.32\textwidth}
		\centering
		\includegraphics[width=\textwidth]{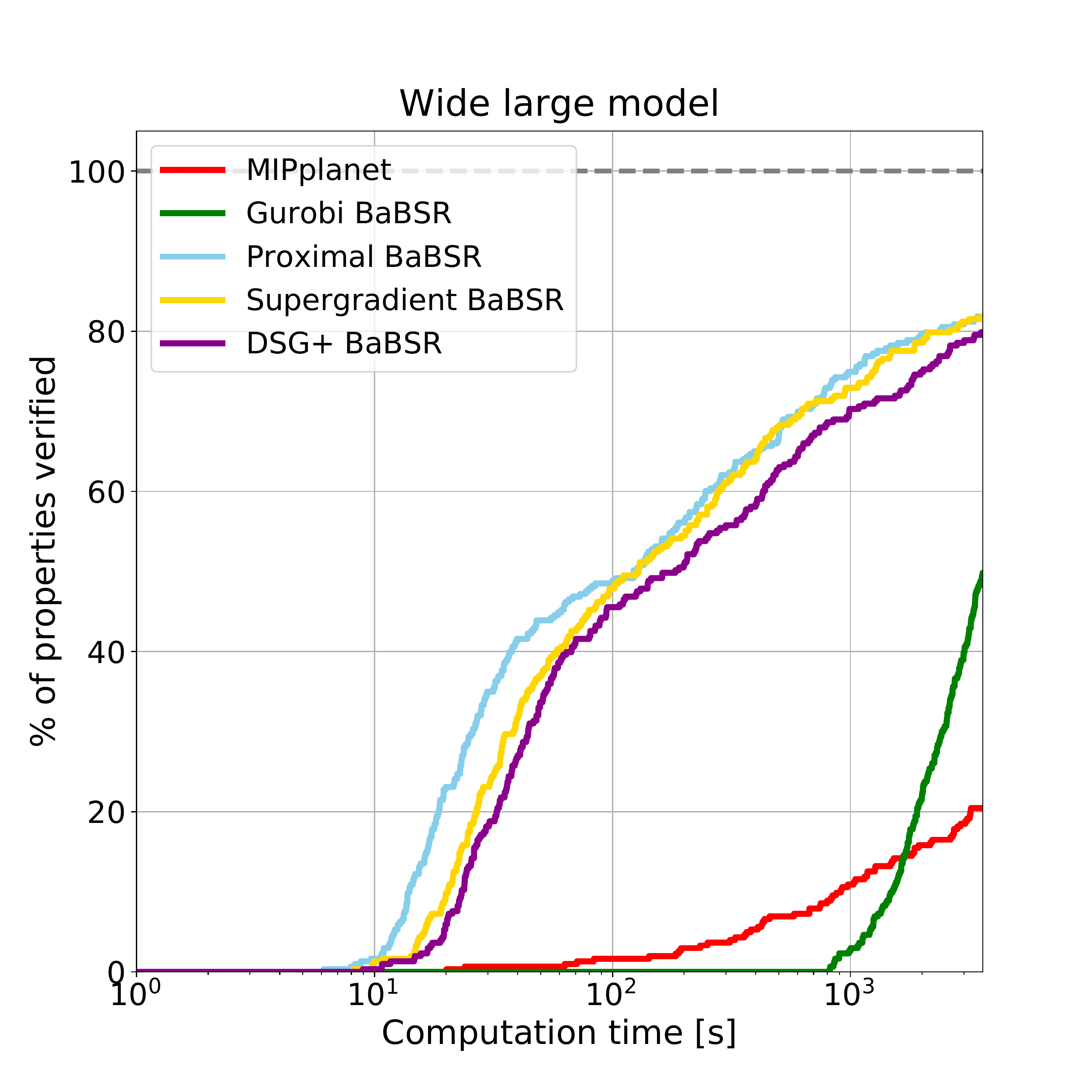}
	\end{subfigure}
	\begin{subfigure}{.32\textwidth}
		\centering
		\includegraphics[width=\textwidth]{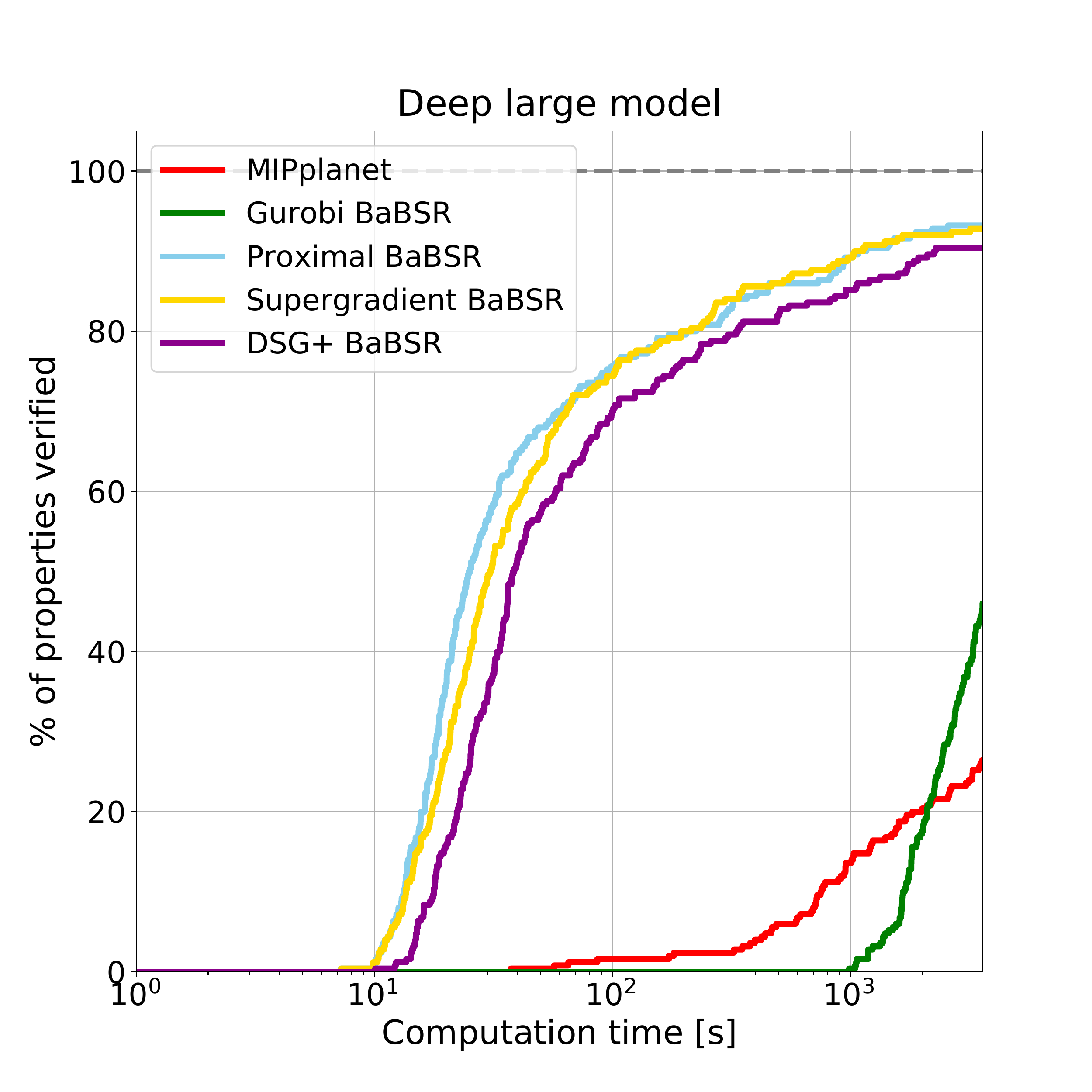}
	\end{subfigure}
	\caption{\vspace{-2pt}Cactus plots for the base, wide and deep models. For each, we compare the bounding methods and complete verification algorithms by plotting the percentage of solved properties as a function of runtime.}
	\label{fig:verification-main}
\end{figure*}

Figure \ref{fig:madry8} presents results as a distribution of runtime and optimality gap. WK by
\citet{Wong2018} performs a single pass over the network per optimization
problem, which allows it to be extremely fast, but this comes at the cost of
generating looser bounds. At the complete opposite end of the spectrum, Gurobi is extremely slow but provides the best achievable bounds. 
Dual iterative methods (DSG+, Supergradient, Proximal) allow the user to choose the
trade-off between tightness and speed. 
In order to perform a fair comparison, we fixed the number of iterations for the various methods so that each of them would take the same average time (see Figure~\ref{fig:madry8-pointwise}). This was done by tuning the iteration ratios on a subset of the images for the results in Figures \ref{fig:madry8}, \ref{fig:madry8-pointwise}.
Note that the Lagrangian Decomposition has a higher cost per iteration due to the more expensive computations related to the more complex primal feasible set. The cost of the proximal method is even larger due to the primal-dual updates and the optimal step size computation.
For DSG+, Supergradient and Proximal, the improved quality of the bounds as compared to the non-iterative method WK shows that there are benefits in actually solving the best relaxation rather than simply evaluating looser bounds. 
Time-limiting Gurobi at the first iteration which exceeded the cost of $400$ Proximal iterations significantly worsens the produced bounds without a comparable cut in runtimes. This is due to the high cost per iteration of the dual simplex algorithm.

By looking at the distributions and at the point-wise comparisons in Figure \ref{fig:madry8-pointwise}, we can see that Supergradient yields consistently better bounds than \textsc{DSG+}. As both methods employ the Adam update rule (and the same hyper-parameters, which are optimal for both), we can conclude that operating on the Lagrangian Decomposition dual \eqref{eq:noncvx_pb} produces better speed-accuracy trade-offs compared to the dual \eqref{eq:dj_prob-main} by \citet{Dvijotham2018}.
This is in line with the expectations from Theorem \ref{theorem}. Furthermore, while a direct application of the Theorem (Dec-DSG+) does improve on the DSG+ bounds, operating on the Decomposition dual \eqref{eq:noncvx_pb} is empirically more effective (see pointwise comparisons in appendix~\ref{sec:experiments-incomplete-appendix}).
Finally, on average, the proximal algorithm yields better bounds than those returned by Supergradient, further improving on the DSG+ baseline. In particular, we stress that the support of optimality gap distribution is larger for Proximal, with a heavier tail towards better bounds.

\subsection{COMPLETE VERIFICATION} \label{sec:experiments-complete}

\sisetup{detect-weight=true,detect-inline-weight=math,detect-mode=true}
\begin{table*}[t!]
	\centering
	\caption{\small For base, wide and deep models, we compare average solving time, average number of solved sub-problems and the percentage of timed out properties. The best performing iterative method is highlighted in bold. \vspace{-2pt}}
	\label{table:models}
	\scriptsize
	\setlength{\tabcolsep}{4pt}
	\aboverulesep = 0.1mm  
	\belowrulesep = 0.2mm  
	\begin{adjustbox}{center}
		\begin{tabular}{
				l
				S[table-format=4.3]
				S[table-format=4.3]
				S[table-format=4.3]
				S[table-format=4.3]
				S[table-format=4.3]
				S[table-format=4.3]
				S[table-format=4.3]
				S[table-format=4.3]
				S[table-format=4.3]
			}
			& \multicolumn{3}{ c }{Base} & \multicolumn{3}{ c }{Wide} & \multicolumn{3}{ c }{Deep} \\
			\toprule
			
			\multicolumn{1}{ c }{Method} &
			\multicolumn{1}{ c }{time(s)} &
			\multicolumn{1}{ c }{sub-problems} &
			\multicolumn{1}{ c }{$\%$Timeout} &
			\multicolumn{1}{ c }{time(s)} &
			\multicolumn{1}{ c }{sub-problems} &
			\multicolumn{1}{ c }{$\%$Timeout} &
			\multicolumn{1}{ c }{time(s)} &
			\multicolumn{1}{ c }{sub-problems} &
			\multicolumn{1}{ c }{$\%$Timeout} \\
			
			\cmidrule(lr){1-1} \cmidrule(lr){2-4} \cmidrule(lr){5-7} \cmidrule(lr){8-10}
			
			\multicolumn{1}{ c }{\textsc{Gurobi BaBSR}}
			&1588.651
			&1342.777
			&10.56
			
			&2912.246
			&630.864
			&50.66
			
			&3007.237
			& 299.913
			&54.00 \\
			
			\multicolumn{1}{ c }{\textsc{MIPplanet}}
			& 2035.565
			& 
			&36.37
			
			& 3118.593
			& 
			&80.00
			
			& 2997.115
			& 
			&73.60 \\
			
			\multicolumn{1}{ c }{\textsc{DSG+ BaBSR}}
			&426.554
			&5312.674
			&6.54
			
			&1048.594
			&4518.116
			&20.00
			
			&509.484
			&1992.345
			&9.60 \\
			
			\multicolumn{1}{ c }{\textsc{Supergradient BaBSR}}
			& 377.104
			& 5079.682
			& \B 5.76
			
			& 920.351
			& 4069.894
			& 18.00
			
			& 390.367
			& 1783.577
			& 7.2 \\
			
			\multicolumn{1}{ c }{\textsc{Proximal BaBSR}}
			&\B 368.876
			&\B 4569.414
			& 5.82
			
			&\B  891.163
			&\B  3343.601
			&\B 18.00
			
			&\B 384.984
			&\B 1655.519
			&\B 6.8 \\
			
			\bottomrule
			
		\end{tabular}
	\end{adjustbox}
\vspace{-10pt}
\end{table*}

We present results for complete verification. In this setting, our goal is to verify whether a network is robust to $\ell_{\infty}$ norm perturbations of radius $\epsilon_{\text{verif}}$. In order to do so, we search for a counter-example (an input point for which the output of the network is not the correct class) by minimizing the difference between the ground truth logit and a randomly chosen logit of images of the CIFAR-10 test set. If the minimum is positive, we have not succeeded in finding a counter-example, and the network is robust.
In contrast to the previous section, we now seek to solve a nonconvex problem like \eqref{eq:noncvx_pb} (where another layer representing the aforementioned difference has been added at the end) directly, rather than an approximation.

In order to perform the minimization exactly,  we employ BaSBR, a Branch and Bound algorithm by \citet{Bunel2020}. In short, Branch and Bound divides the verification domain into a number of smaller problems, for which it repeatedly computes upper and lower bounds. At every iteration, BaSBR picks the sub-problem with the lowest lower bound and creates two new sub-problems by fixing the most ``impactful" ReLU to be passing or blocking. The impact of a ReLU is determined by estimating the change in the sub-problem's output lower bound caused by making the ReLU non-ambiguous. 
Sub-problems which cannot contain the global lower bound are progressively discarded. 

The computational bottleneck of Branch and Bound is the lower bounds computation, for which we will employ a subset of the methods in section \ref{sec:experiments-setting}.
Specifically, we want to compare the performance of DSG+, Supergradient and Proximal with Gurobi (as employed in \citet{Bunel2020}). For this purpose, we run the various Branch and Bound implementations on the dataset employed by \citet{Lu2020} to test their novel Branch and Bound splitting strategy. The dataset picks a non-correct class and a perturbation radius $\epsilon_{\text{verif}}$ for a subset of the CIFAR-10 test images, and runs Branch and Bound on three different convolutional network architectures of varying size: a ``base" network, a ``wide" network, and a ``deep" network. Further details on the dataset and architectures are provided in appendix~\ref{sec:experiments-complete-appendix}.
 
We compare the Branch and Bound implementations with MIPplanet to provide an additional verification baseline. MIPplanet computes the global lower bound by using Gurobi to solve the Mixed-Integer Linear problem arising from the Planet relaxation \citep{Bunel2020,Ehlers2017}.  
As explained in section \ref{sec:implementation}, due to the highly parallelizable nature of the dual iterative algorithms, we are able to compute lower bounds for multiple sub-problems at once for DSG+, Supergradient and Proximal, whereas the domains are solved sequentially for Gurobi. The number of simultaneously solved sub-problems is $300$ for the base network, and $200$ for the wide and deep networks. Intermediate bound computations (see $\S$\ref{sec:experiments-setting}) are performed in parallel as well, with smaller  batch sizes to account for the larger width of intermediate layers (over which we batch as well).
For both supergradient methods (our Supergradient, and DSG+), we decrease the step size linearly from $\alpha=10^{-3}$ to $\alpha=10^{-4}$, while for Proximal, we do not employ momentum and keep the weight of the proximal terms fixed to $\eta=10^2$ for all layers throughout the iterations. As in the previous section, the number of iterations for the bounding algorithms are tuned to employ roughly the same time: we use $100$ iterations for Proximal, $160$ for Supergradient, and $260$ for DSG+. For all three, the dual variables are initialized from the dual solution of the lower bound computation of the parent node in the Branch and Bound tree. We time-limit the experiments at one hour.

Consistently with the incomplete verification results in the last section, Figure \ref{fig:verification-main} and Table \ref{table:models} show that the Supergradient overperforms DSG+, confirming the benefits of our Lagrangian Decomposition approach. Better bounds decrease the number of sub-problems that Branch and Bound needs to solve, reducing the overall verification time. Furthermore, Proximal provides an additional increase in performance over DSG+, which is visible especially over the properties that are easier to verify. The gap between competing bounding methods increases with the size of the employed network, both in terms of layer width and network depth: at least an additional $2\%$ of the properties times out when using DSG+ on the larger networks. A more detailed experimental analysis of the base model data is presented in appendix \ref{sec:experiments-complete-appendix}. 

\section{DISCUSSION}
We have presented a novel dual approach to compute bounds over the activation of
neural networks based on Lagrangian Decomposition. It provides significant benefits compared to off-the-shelf
solvers and improves on both looser relaxations and on a previous method based on Lagrangian relaxation. 
As future work, it remains to investigate whether better complete  verification results can be obtained by combining our supergradient and proximal methods.
Furthermore, it
would be interesting to see if similar derivations could be used to
make tighter but more expensive relaxations computationally feasible. Improving
the quality of bounds may allow the training of robust networks to avoid
over-regularisation.

\subsubsection*{Acknowledgments}
ADP was supported by the EPSRC Centre for Doctoral Training in Autonomous Intelligent Machines and Systems, grant EP/L015987/1.
RB wishes to thank Leonard Berrada for helpful discussions on the convex hull of sigmoid activation functions.

\bibliographystyle{apalike}
\bibliography{standardstrings,oval,decomposition_verification}

\clearpage

\iftoggle{appendix}{
	\onecolumn  
	\appendix

	\section{Sigmoid Activation function}
\label{sec:sigmoid_act}
This section describes the computation highlighted in the paper in the context
where the activation function $\sig{x}$ is the sigmoid function:
\begin{equation}
  \sig{x} = \frac{1}{1+e^{-x}}
\end{equation}
A similar methodology to the ones described in this section could be used to
adapt the method to work with other activation function such as hyperbolic
tangent, but we won't discuss it.

We will start with a reminders about some properties of the sigmoid activation
function. It takes values between 0 and 1, with $\sig{0}=0.5$. We can easily
compute its derivatives:
\begin{equation}
  \begin{split}
    \sigma^\prime\left( x \right) &= \sig{x} \times (1 - \sig{x})\\
    \sigma^{\prime\prime}\left( x \right) &= \sig{x} \times (1 - \sig{x}) \times \left( 1 - 2 \sig{x} \right)\\
  \end{split}
\end{equation}

If we limit the domain of study to the negative inputs ($\left[ -\infty, 0
\right]$), then the function $x \mapsto \sig{x}$ is a convex function, as can
be seen by looking at the sign of the second derivative over that domain.
Similarly, if we limit the domain of study to the positive inputs ($\left[ 0,
  \infty \right]$), the function is concave.

\begin{figure}
  \begin{subfigure}{.3\textwidth}
    \resizebox{\textwidth}{!}{\input{figures/sigmoid_fig/blocking.pgf}}
    \caption{\label{fig:blocked_sigmoid}The upper bound is in Case 1, while the
      lower bound is in the special case of Case 2 with only one piece.}
  \end{subfigure}%
  \hfill%
  \begin{subfigure}{.3\textwidth}
    \resizebox{\textwidth}{!}{\input{figures/sigmoid_fig/passing.pgf}}
    \caption{\label{fig:passing_sigmoid}The upper bound is in the special case
      of Case 2 with only one piece, while the lower bound is in Case 1.}
  \end{subfigure}%
  \hfill%
  \begin{subfigure}{.3\textwidth}
    \resizebox{\textwidth}{!}{\input{figures/sigmoid_fig/ambiguous.pgf}}
    \caption{\label{fig:amb_sigmoid}Both upper and lower bounds are in Case 2
      with the two pieces visible.}
  \end{subfigure}
  \caption{\label{fig:sigmoid_cvx_hull}Convex hull of the Sigmoid activation
    function for different input bound configurations.}
\vspace{20pt}
\end{figure}

\subsection{Convex hull computation}
In the context of ReLU, the convex hull of the activation function is given by
equation\eqref{eq:planet-relu-hull}, as introduced by \citet{Ehlers2017}. We will
now derive it for sigmoid functions. Upper and lower bounds will be dealt in the
same way, so our description will only focus on how to obtain the concave upper
bound, limiting the activation function convex hull by above. The computation
to derive the convex lower bound is equivalent.

Depending on the range of inputs over which the convex hull is taken, the form
of the concave upper bound will change. We distinguish two cases.

\paragraph{Case 1: $\sigma^\prime\left(\ub_k\right) \geq \ulslope$.}
We will prove that in this case, the upper bound will be the line passing through the points $(\lb_k,
\sig{\lb_k})$ and $(\ub_k, \sig{\lb_k})$. The equation of it is given by:
\begin{equation}
  \phiub(\xb) = \ulslope \left( \xb - \lb_k \right) + \sig{\lb_k}
  \label{eq:sig_lin_upper_bound}
\end{equation}

Consider the function $d(\xb) = \phiub(\xb) - \sig{\xb}$. To show that $\phiub$
is a valid upper bound, we need to prove that $\forall \xb \in \left[ \lb_k,
  \ub_k \right],\ d(\xb) \geq 0$. We know that $d(\lb_k) = 0$ and $d(\ub_k) =
0$, and that $d$ is a continuous function. Its derivative is given by:
\begin{equation}
  d^\prime(\xb) = \ulslope - \sig{\xb} (1 - \sig{\xb}).
\end{equation}
To find the roots of $d^\prime$, we can solve $d^\prime(\xb) = 0$ for the value
of $\sig{\xb}$ and then use the logit function to recover the value of $\xb$. In
that case, this is only a second order polynomial, so it can admit at most two
roots.

We know that $\lim_{\xb \to \infty} d^\prime(\xb) \geq 0$, and our hypothesis
tells us that $d^\prime(\ub_k) \leq 0$. This means that at least one of the root
lies necessarily beyond $\ub_k$ and therefore, the derivative of $d$ change
signs at most once on the $\left[ \lb_k, \ub_k \right]$ interval. If it never
changes sign, it is monotonous. Given that the values taken at both extreme
points are the same, $d$ being monotonous would imply that $d$ is constant,
which is impossible. We therefore conclude that this means that the derivative
change its sign exactly once on the interval, and is therefore unimodal. As we know
that $d^\prime(\ub_k) \leq 0$, this indicates that $d$ is first increasing and
then decreasing. As both extreme points have a value of zero, this means that
$\forall \xb \in \left[ \lb_k, \ub_k \right], \ d(x) \geq 0$.

From this result, we deduce that $\phiub$ is a valid upper bound of $\sigma$. As a
linear function, it is concave by definition. Given that it constitutes the line
between two points on the curve, all of its points necessarily belong to the convex
hull. Therefore, it is not possible for a concave upper bound of the activation
function to have lower values. This means that $\phiub$ defines the upper
bounding part of the convex hull of the activation function.

\paragraph{Case 2: $\sigma^\prime \left( \ub_k \right) \leq \ulslope$.}
In this case, we will have to decompose the upper bound into two parts, defined
as follows:
\begin{subnumcases}{\label{eq:sig_pw_upper_bound}\phiub(\xb)}
  \tlslope\left( \xb - \lb_k \right) +
  \sig{\lb_k}  & if $\xb \in \left[ \lb_k, \tb_k \right]$\\
  \sig{\xb} & if $\xb \in \left[ \tb_k, \ub_k \right]$\label{eq:ub_match_fun},
\end{subnumcases}
where $\tb_k$ is defined as the point such that $\sigma^\prime\left( \tb_k
\right) = \tlslope$ and $\tb_k > 0$. The value of $\tb_k$ can be computed by
solving the equation $\sig{\tb_k}(1 - \sig{\tb_k}) = \tlslope$, which can be
done using the Newton-Raphson method or a binary search. Note that this needs to
be done only when defining the problem, and not at every iteration of the
solver. In addition, the value of $\tb_k$ is dependant only on $\lb_k$ so it's
possible to start by building a table of the results at the desired accuracy and
cache them.

Evaluating both pieces of the function of equation\eqref{eq:sig_pw_upper_bound}
at $\tb_k$ show that $\phiub$ is continuous. Both pieces are concave (for $\xb
\geq \tb_k \geq 0$, $\sigma$ is concave) and they share a supergradient (the
linear function of slope $\tlslope$) in $\tb_k$, so $\phiub$ is a concave
function. The proof we did for {\bf Case 1} can be duplicated to show that the
linear component is the best concave upper bound that can be achieved over the
interval $\left[ \lb_k, \tb_k \right]$. On the interval $\left[ \tb_k, \ub_k
\right]$, $\phiub$ is equal to the activation function, so it is also an upper
bound which can't be improved upon. Therefore, $\phiub$ is the upper bounding
part of the convex hull of the activation function.

Note that a special case of this happens when $\lb_k \geq \tb_k$. In which case,
$\phiub$ consists of only equation~\eqref{eq:ub_match_fun}.

All cases are illustrated in Figure~\ref{fig:sigmoid_cvx_hull}. Case 1 is shown
in \ref{fig:blocked_sigmoid}, where the upper bound contains only the linear
upper bound. Case 2 with both segments is visible in
Figure\ref{fig:amb_sigmoid}, with the cutoff points $tb_k$ highlighted by a
green dot, and the special case with $\lb_k \geq \tb_k$ is demonstrated in
Figure~\ref{fig:passing_sigmoid}.

\subsection{Solving the $\mathcal{P}_k$ subproblems over sigmoid activation}

As a reminder, the problem that needs to be solved is the following (where $\mbf{c}_{k+1} = -\lambdabhat_{k+1}$, $\mbf{c}_{k} = \lambdabhat_{k}$):
\begin{equation}
  \begin{split}
    [\xbhat_{B, k}, \xbhat_{A, k+1}] = \argmin_{\zbhat_{B, k}, \zbhat_{A, k+1}}& \quad
    \mbf{c}_k^T \zbhat_{B, k} + \mbf{c}_{k+1}^T \zbhat_{A, k+1}\\
    \text{s.t } & \quad \lb_{k} \leq \zbhat_{B, k} \leq \ub_{k}\\
    & \quad \texttt{cvx\_hull}_\sigma\left( \zbhat_{B,k}, \zb_k, \lb_k, \ub_k\right)\\
    & \quad \zbhat_{A, k+1} = W_{k+1} \zb_{k} + \mathbf{b}_{k+1},
  \end{split}
  \label{eq:pk_sub_for_sig}
\end{equation}
where $\texttt{cvx\_hull}$ is defined either by
equations~\eqref{eq:sig_lin_upper_bound} or \eqref{eq:sig_pw_upper_bound}.
In this case, we will still be able to compute a closed form solution.

We will denote $\phiub$ and $\psiub$ the upper and lower bound functions
defining the convex hull of the sigmoid function, which can be obtained as
described in the previous subsection. If we use the last equality constraint of
Problem~\eqref{eq:pk_sub_for_sig} to replace the $\zbhat_{A, k+1}$ term in the objective
function, we obtain  \mbox{$\mbf{c}_k^T \zbhat_{B, k} + \mbf{c}_{k+1}^T  W_{k+1}
  \zb_{k}$}. Depending on the sign of $\mbf{c}_{k+1}^T  W_{k+1}$, $\zb_k$ will
either take the value $\phiub$ or $\psiub$, resulting in the following problem:

\begin{equation}
  \begin{split}
    \min_{\zbhat_{B,k}} \quad & \mbf{c}^T_{k} \zbhat_{B,k} + \left[ \mbf{c}^T_{k+1}W_{k+1} \right]_{-} \phiub\left(\zbhat_{B,k}\right) + \left[ \mbf{c}^T_{k+1}W_{k+1} \right]_{+} \psiub\left( \zbhat_{B, k}\right)\\
    \text{s.t } & \quad \lb_{k} \leq \zbhat_{B, k} \leq \ub_{k}.
    \label{eq:siglinopt}
  \end{split}
\end{equation}

To solve this problem, several observations can be made: First of all, at that
point, the optimisation decomposes completely over the components of
$\zbhat_{B,k}$ so all problems can be solved independently from each other.
The second observation is that to solve this problem, we can decompose the minimisation over the
whole range $\left[ \lb_k, \ub_k \right]$ into a set of minimisation over
the separate pieces, and then returning the minimum corresponding to the piece
producing the lowest value.

The minimisation over the pieces can be of two forms. Both bounds can be linear
(such as between the green dotted lines in Figure~\ref{fig:amb_sigmoid}), in
which case the problem is easy to solve by looking at the signs of the
coefficient of the objective function. The other option is that one of the
bound will be equal to the activation function (such as in
Figures~\ref{fig:blocked_sigmoid} or \ref{fig:passing_sigmoid}, or in the outer
sections of Figure~\ref{fig:amb_sigmoid}), leaving us with a problem of the
form:
\begin{equation}
  \begin{split}
    \min_{\zbhat_{B, k}} \quad & \mbf{c}_\text{lin}^T \zbhat_{B, k} + \mbf{c}_{\sigma}^T \sig{\zbhat_{B,k}}\\
    & \lb \leq \zbhat_{B,k} \leq \ub,
  \end{split}
  \label{eq:with_sig_subp}
\end{equation}
where $\lb$, $\ub$, $\mbf{c}_\text{lin}$ and $\mbf{c}_\sigma$ will depend on
what part of the problem we are trying to solve.

This is a convex problem so the value will be reached either at the extremum of
the considered domain ($\lb$ or $\ub$), or it will be reached at the points
where the derivative of the objective functions cancels. This corresponds to
the roots of $\mbf{c}_\text{lin} + \mbf{c}_{\sigma}^T \sig{\zbhat_{B,k}}(1 - \sig{\zbhat_{B,k}})$.
Provided that $1 + \frac{4 \mbf{c}_\text{lin}}{\mbf{c}_\sigma} \geq 0$, the
possible roots will be given by \mbox{$\sigma^{-1}\left( \frac{1 \pm \sqrt{1 + \frac{4
        \mbf{c}_\text{lin}}{\mbf{c}_\sigma}}}{2} \right)$}, with $\sigma^{-1}$
being the logit function, the inverse of the sigmoid function $\left(
  \sigma^{-1}(\xb) = \log\left( \frac{\xb}{1-\xb} \right) \right)$. To solve
problem~\eqref{eq:with_sig_subp}, we evaluate its objective function at the
extreme points ($\lb$ and $\ub$) and at those roots if they are in the feasible
domain, and return the point corresponding to the minimum score achieved. With
this method, we can solve the $\mathcal{P}_k$ subproblems even when the
activation function is a sigmoid.


	\section{Intermediate bounds} \label{sec:intermediate}
	
	Intermediate bounds are obtained by solving a relaxation of \eqref{eq:noncvx_pb}
	over subsets of the network. Instead of defining the objective
	function on the activation of layer $n$, we define it over $\zbhat_k$, iteratively for $k \in [1..n-1]$. Computing intermediate
	bounds with the same relaxation would be the computational cost of this approach: we need to solve two
	LPs for each of the neurons in the network, and even if (differently from off-the-shelf LP solvers) our method allows for easy parallelization within each layer, the layers need to be tackled sequentially.
	Therefore, depending on the computational budget, the employed relaxation for intermediate bounds might be looser than \eqref{eq:planet-relax} or \eqref{eq:dj_prob-main}: for instance, the one by \citet{Wong2018}.
	
	\section{Implementation details for Proximal method} \label{sec:prox-appendix}
We provide additional details for the implementation of the proximal method. We start with the optimal step size computation, define our block-coordinate updates for the primal variables, and finally provide some insight on acceleration.
\subsection{Optimal Step Size Computation}
Having obtained the conditional gradients $\xbhat_k\ \forall k \in [0 .. n-1]$ for our problem, we need to decide a
step size. Computing the optimal step size amounts to solving a one dimensional
quadratic problem:
\begin{equation}
\gamma^* = \argmin_{\gamma \in \left[ 0, 1 \right]}\mathcal{L}(\gamma \xbhat + (1 - \gamma)\zbhat, \lambdabhat).
\end{equation}
Computing the gradient with regards to $\gamma$ and setting it to 0 will return
the optimal step size.  Denoting the gradient of the Augmented Lagrangian~\eqref{eq:z_updates} with respect to $(\zbhat_{B,k} -\zbhat_{A,k})$ as \mbox{$\nabla_{(\zbhat_{B,k} -\zbhat_{A,k})} \mathcal{L}\left( \zbhat,\lambdabhat\right) = \lambdabhat_k + \frac{\zbhat_{B,k} - \zbhat_{A,k}}{\eta_k} = \mbf{g}_k$}, this gives us the following formula:
\begin{equation}
\gamma^* = \texttt{clamp}_{[0, 1]}\left(-\frac{
	\left(\xbhat_{A, n} - \zbhat_{A, n}  \right)
	+ \sum_k \mbf{g}_k^T \left( \xbhat_{B,k} - \zbhat_{B,k} - \xbhat_{A,k} + \zbhat_{A,k} \right)
}{
	\sum_k \frac{1}{\eta_k} \|\xbhat_{B,k} - \zbhat_{B,k} - \xbhat_{A,k} + \zbhat_{A,k} \|^2
}\right)
\label{eq:optimal_gamma}
\end{equation}


\subsection{Gauss-Seidel style updates}

At iteration $t$, the Frank-Wolfe step takes the following form:
\begin{equation}
\zbhat^{t+1} = \gamma^* \xb + (1 - \gamma^*) \zbhat^t 
\label{eq:jacobi-update}
\end{equation}
where the update is performed on the variables associated to all the network neurons at once. 

As the Augmented Lagrangian \eqref{eq:z_updates} is smooth in the primal variables, we can take a conditional gradient step after each $\xbhat_k$ computation, resulting in Gauss-Seidel style updates (as opposed to Jacobi style updates \ref{eq:jacobi-update})~\citep{Bertsekas1989}. As the values of the primals at following layers are inter-dependent through the gradient of the Augmented Lagrangian, these block-coordinate updates will result in faster convergence.
The updates become:
\begin{equation}
\zbhat_{k}^{t+1} = \gamma_k^* \xbhat_k + (1 - \gamma_k^*) \zbhat_{k}^t 
\end{equation}
where the optimal step size is given by:
\begin{equation}
\gamma_{k} ^*= \texttt{clamp}_{[0, 1]}\left(-\frac{
	\mbf{g}_k^T \left( \xbhat_{B,k} - \zbhat_{B,k})  + \mbf{g}_{k+1}^T (\zbhat_{A,k+1} - \xbhat_{A,k+1}) \right)
}{
	\frac{1}{\eta_k} \|\xbhat_{B,k} - \zbhat_{B,k} \|^2 + \frac{1}{\eta_{k+1}}  \|\zbhat_{A,k+1} - \xbhat_{A,k+1}  \|^2
}\right)
\label{eq:optimal-gamma-gauss}
\end{equation}
with the special cases $\xbhat_{B,0} = \zbhat_{B,0} = 0$ and:
\begin{equation}
\gamma_{n-1}^* = \texttt{clamp}_{[0, 1]}\left(-\frac{
	\mbf{g}_{n-1}^T \left( \xbhat_{B,n-1} - \zbhat_{B,n-1})  + (\xbhat_{A, n} - \zbhat_{A, n}) \right)
}{
	\frac{1}{\eta_{n-1}} \|\xbhat_{B,n-1} - \zbhat_{B,n-1} \|^2 
}\right)
\label{eq:optimal-gamma-gauss-final}
\end{equation}

\subsection{Momentum}
The supergradient methods that we implemented both for our dual \eqref{eq:nonprox_problem} and for \eqref{eq:dj_prob-main} by \citet{Dvijotham2018} rely on Adam~\citep{Kingma2015} to speed-up convergence. 
Inspired by its presence in Adam, and by the work on accelerating proximal methods \cite{Lin2017,Salzo12}, we apply momentum on the proximal updates.

By closely looking at equation \eqref{eq:lambda_updates}, we can draw a similarity between the dual update in the proximal algorithm and supergradient descent. The difference is that the former operation takes place after a closed-form minimization of the linear inner problem in $\zbhat_A, \zbhat_B$, whereas the latter after some steps of an optimization algorithm that solves the quadratic form of the Augmented Lagrangian \eqref{eq:z_updates}. We denote the (approximate) argmin of the Lagrangian at the $t$-th iteration of the proximal method (method of multipliers) by $\zbhat^{t, \dagger}$.

Thanks to the aforementioned similarity, we can keep an exponential average of the gradient-like terms with parameter $\mu \in [0,1]$ and adopt momentum for the dual update, yielding:
\begin{equation}
\begin{split}
\pib_k^{t+1} &= \mu \pib_k^{t} + \frac{\zbhat^{t, \dagger}_{B,k} - \zbhat^{t, \dagger}_{A,k}}{\eta_k} \\
\lambdabhat_k^{t+1} &= \lambdabhat_k^{t} + \pib_k^{t+1} 
\end{split}
\label{eq:momentum_updates}
\end{equation}

As, empirically, a decaying step size has a positive effect on Adam, we adopt the same strategy for the proximal algorithm as well, resulting in an increasing $\eta_k$.
The augmented Lagrangian then becomes:
\begin{equation}
\begin{split}
\left[ \zb^{t}, \zbhat^{t} \right] \ = \smash{\argmin_{\zb, \zbhat}}\, \mathcal{L}\left( \zbhat, \lambdabhat^t\right)&
\quad \text{s.t. } \mathcal{P}_0(\zb_0, \zbhat_{A,1}); \quad \mathcal{P}_k(\zbhat_{B, k}, \zbhat_{A, k+1})\ \quad k \in \left[1..n-1\right]\\
\text{where } \auglag=\ & \Bigg[ \zbhat_{A, n} + \sum_{k=1..n-1} \lambdabhat_k^T \left( \zbhat_{B, k} - \zbhat_{A, k} \right) + \sum_{k=1..n-1} \frac{1}{2\eta_k}  \|\zbhat_{B,k} - \zbhat_{A,k} \|^2\\
  &\ \ - \sum_{k=1..n-1} \quad \frac{\eta_k}{2}  \| \mu \pib_k \|^2 \Bigg].
\end{split}
\label{eq:new_ag}
\end{equation}
We point out that the Augmented Lagrangian's gradient \mbox{$\nabla_{(\zbhat_{B,k} -\zbhat_{A,k})} \mathcal{L}\left( \zbhat,\lambdabhat\right)$} remains unvaried with respect to the momentum-less solver hence, in practice, the main change is the dual update formula \eqref{eq:momentum_updates}.

	\section{Algorithm summary}
\label{sec:algo_summary}
We provide high level summaries of our solvers: supergradient (Algorithm \ref{alg:subg}) and proximal (Algorithm \ref{alg:prox}).

For both methods, we decompose problem~\eqref{eq:planet-relax} into several subset of constraints, duplicate the variables to make the subproblems independent, and enforce their consistency by introducing dual variables $\lambdabhat$, which
we are going to optimize. This results in problem~\eqref{eq:nonprox_problem}, which we solve using one of the two algorithms.
Our initial dual solution is given by the algorithm of
\citet{Wong2018}. The details can be found in Appendix~\ref{sec:WK_init}.

\begin{minipage}{.90\textwidth}
	\begin{algorithm}[H]
		\caption{Supergradient method}\label{alg:subg}
		\begin{algorithmic}[1]
			\algrenewcommand\algorithmicindent{1.0em}%
			\Function{subg\_compute\_bounds}{$\{W_k, \mbf{b}_k, \mbf{l}_k, \mbf{u}_k\}_{k=1..n}$}
			\State Initialise dual variables $\lambdabhat$ using the algo of \citet{Wong2018}
			\For{$\mtt{nb\_iterations}$}\label{alg:inner_loopstart}
			\State $\zbhat^* \leftarrow$ inner minimization using $\S$\ref{sec:innerminp0} and $\S$\ref{sec:innerminpk} \label{alg:inner_min}
			\State Compute supergradient using $\nabla_{\lambdabhat} q(\lambdabhat^t) = \zbhat_B^* - \zbhat_A^* $ 
			\State $\lambdabhat^{t+1} \leftarrow$ Adam's update rule \cite{Kingma2015} \label{alg:adam-update}
			\EndFor\label{alg:inner_loopend}
			\State\Return $q(\lambdabhat)$
			\EndFunction
		\end{algorithmic}
	\end{algorithm}
\end{minipage}%

The structure of our employed supergradient method is relatively simple. We iteratively minimize over the primal variables (line \ref{alg:inner_min}) in order to compute the supergradient, which is then used to update the dual variables through the Adam update (line \ref{alg:adam-update}).

\begin{minipage}{.90\textwidth}
  \begin{algorithm}[H]
    \caption{Proximal method}\label{alg:prox}
    \begin{algorithmic}[1]
      \algrenewcommand\algorithmicindent{1.0em}%
      \Function{prox\_compute\_bounds}{$\{W_k, b_k, l_k, u_k\}_{k=1..n}$}
      \State Initialise dual variables $\lambdabhat$ using the algo of \citet{Wong2018}\label{alg:init_dual}
      \State Initialise primal variables $\zbhat$ at the conditional gradient of $\lambdabhat$
        \For{$\mtt{nb\_outer\_iterations}$}\label{alg:outer_loop}
          \State Update dual variables using equation~\eqref{eq:lambda_updates} or \eqref{eq:momentum_updates}\label{alg:lambda_step}
          \For{$\mtt{nb\_inner\_iterations}$}\label{alg:inner_loopstart}
          	\For{$k \in [0..n-1]$} \label{alg:layer-iter}
	            \State Compute layer conditional gradient $[\xbhat_k]$ using using $\S$\ref{sec:innerminp0} and $\S$\ref{sec:innerminpk} \label{alg:cond_grad}
	            \State Compute layer optimal step size $\gamma_k^*$ using \eqref{eq:optimal_gamma}\label{alg:opt_stepsize}
	            \State $\begin{array}{l}\hspace{-.5em}\text{Update layer primal variables:}\\
	                      \qquad \zbhat_k = \gamma_k^* \xbhat_k + (1 - \gamma_k^*) \zbhat_k\\
	                    \end{array}$\label{alg:fw_step}
	        \EndFor
          \EndFor\label{alg:inner_loopend}
        \EndFor
      \State\Return $q(\lambdabhat)$
      \EndFunction
    \end{algorithmic}
  \end{algorithm}
\end{minipage}%

The proximal method, instead alternates updates to the dual variables (applying the updates from
\eqref{eq:lambda_updates}; line \ref{alg:lambda_step}) and to
the primal variables (update of \eqref{eq:z_updates}; lines \ref{alg:inner_loopstart} to
\ref{alg:inner_loopend}). The inner minimization problem is solved only
approximately, using the Frank-Wolfe algorithm (line \ref{alg:fw_step}), applied in a block-coordinate fashion over the network layers (line \ref{alg:layer-iter}).
We have closed form solution for both the conditional gradient (line
\ref{alg:cond_grad}) and the optimal step-size (line \ref{alg:opt_stepsize}).


	\vspace{30pt}
\section{Comparison between the duals}
\label{sec:dual_comp_proof}

We will show the relation between the decomposition done by
\citet{Dvijotham2018} and the one proposed in this paper. We will show that, in
the context of ReLU activation functions, from any dual solution of their dual
providing a bound, our dual provides a better bound. 

\begin{theorem*}
	Let us assume $\sigma(\zbhat_k) = \max(\mathbf{0}, \zbhat_k)$. For any solution $\mub, \lambdab$ of dual \eqref{eq:dj_prob-main} by \citet{Dvijotham2018} yielding bound $d(\mub, \lambdab)$, it holds that $q(\mub) \geq d(\mub, \lambdab)$.
\end{theorem*}

\begin{proof}

We start by reproducing the formulation that \citet{Dvijotham2018} uses, which
we slightly modify in order to have a notation more comparable to
ours\footnote{Our activation function $\sigma$ is denoted $h$ in their paper.
  The equivalent of $\xb$ in their paper is $\zb$ in ours, while the equivalent
  of their $\zb$ is $\zbhat$. Also note that their paper use the computation of
  upper bounds as examples while ours use the computation of lower bounds.}.
With our convention, doing the decomposition to obtain Problem (6) in
\citep{Dvijotham2018} would give:
\begin{equation}
  \begin{split}
    \min_{\zb, \zbhat} & \quad W_n\zb_{n-1} + \mbf{b}_n
    + \sum_{k=1}^{n-1} \mub_k^T \left( \zbhat_k - W_k \zb_{k-1} - \mbf{b}_k \right)
    + \sum_{k=1}^{n-1} \lambdab_k^T \left( \zb_k - \sigma(\zbhat_k) \right)\\
    \text{such that }\quad& \lb_k \leq \zbhat_k \leq \ub_k \qquad\qquad \text{for } k \in \left[1,n-1\right]\\
    & \sigma(\lb_k) \leq \zb_k \leq \sigma(\ub_k) \qquad \text{for } k \in \left[1,n-1\right]\\
    & \zb_0 \in \mathcal{C}
  \end{split}
  \label{eq:dj_prob}
\end{equation}

Decomposing it, in the same way that \citet{Dvijotham2018} do it in order to
obtain their equation (7), we obtain
\begin{equation}
  \begin{split}
    & \mbf{b}_n - \sum_{k=1}^{n-1} \mub_k^T \mbf{b}_k
    + \sum_{k=1}^{n-1}\min_{\zbhat_k \in [\lb_k, \ub_k]}\left( \mub_k^T \zbhat_k - \lambdab_k^T\sigma(\zbhat_k) \right)\\
    &\qquad + \sum_{k=1}^{n-1}\min_{\zb_k \in [\sig{\lb_k}, \sig{\ub_k}]}\left( -\mub_{k+1}^TW_{k+1} \zb_k + \lambdab_k^T\zb_k \right)
     + \min_{\zb_0 \in \mathcal{C}} -\mub_1^T W_1 \zb_0.
  \end{split}
  \label{eq:dj_dual_decomp}
\end{equation}
(with the convention that $\mub_n = -I$)

As a reminder, the formulation of our dual \eqref{eq:nonprox_problem} is:
\begin{equation}
  \begin{split}
    q\left( \lambdabhat \right) = \text{min }_{\zb, \zbhat}\  & \zbhat_{A, n}
    + \Sigma_{k=1..n-1} \lambdabhat_k^T \left( \zbhat_{B, k} - \zbhat_{A, k} \right)\\
    \text{s.t. }\quad & \mathcal{P}_0(\zb_0, \zbhat_{A,1}); \quad \mathcal{P}_k(\zbhat_{B, k}, \zbhat_{A, k+1}) \qquad \text{for } k \in \left[1..n-1\right]\\
  \end{split}
\end{equation}
which can be decomposed as:
\begin{equation}
  \begin{split}
  q\left( \lambdabhat \right) = \sum_{k=1}^{n-1}&\left(
    \min_{\begin{array}{c}\zbhat_{B, k}, \zbhat_{A, k+1} \\ \in \mathcal{P}_k\left( \cdot, \cdot \right)\end{array}}
    \lambdabhat_k^T \zbhat_{B,k}\ -\ \lambdabhat_{k+1}^T \zbhat_{A, k+1}
  \right)
  + \min_{\begin{array}{c}\zb_0, \zbhat_{A,1}\\\in \mathcal{P}_0\left( \cdot, \cdot \right)\end{array}} -\lambdabhat_1^T \zbhat_{A, 1},
  \end{split}
  \label{eq:dual_decomp_vs_dj}
\end{equation}
with the convention that $\lambdabhat_n = -I$. We will show that when we chose the dual variables such that
\begin{equation}
  \lambdabhat_k = \mub_k,
  \label{eq:vs_dj_replacement}
\end{equation}
we obtain a tighter bound than the ones given by \eqref{eq:dj_dual_decomp}.

We will start by showing that the term being optimised over $\mathcal{P_0}$ is
equivalent to some of the terms in \eqref{eq:dj_dual_decomp}:
\begin{equation}
  \min_{\mathclap{\begin{array}{c}\zb_0, \zbhat_{A,1}\\\in \mathcal{P}_0\left( \cdot, \cdot \right)\end{array}}}
  -\lambdabhat_1^T \zbhat_{A, 1}
  = \min_{\mathclap{\begin{array}{c}\zb_0, \zbhat_{A,1}\\\in \mathcal{P}_0\left( \cdot, \cdot \right)\end{array}}}
  - \mub_1^T \zbhat_{A, 1}
  = -\mub_1^T \mbf{b}_1 + \min_{\zb_0 \in \mathcal{C}} - \mub_1^T W_1 \zb_0
  \label{eq:pb0_lb}
\end{equation}
The first equality is given by the replacement
formula~\eqref{eq:vs_dj_replacement}, while the second one is given by
performing the replacement of $\zbhat_{A, 1}$ with his expression in $\mathcal{P}_0$

We will now obtain a lower bound of the term being optimised over
$\mathcal{P}_{k}$. Let's start by plugging in the values using the
formula~\eqref{eq:vs_dj_replacement} and replace the value of $\zbhat_{A, k+1}$
using the constraint.
\begin{equation}
    \min_{\mathclap{\begin{array}{c}\zbhat_{B, k}, \zbhat_{A, k+1} \\ \in \mathcal{P}_k\left( \cdot, \cdot \right)\end{array}}}
    \left( \lambdabhat_k^T \zbhat_{B,k}\ -\ \lambdabhat_{k+1}^T \zbhat_{A, k+1} \right)
    = -\mub_{k+1}^T \mbf{b}_{k+1}
    + \qquad \min_{\mathclap{\begin{array}{l}
                             \qquad\zbhat_{B, k} \in [\lb_k, \ub_k]\\
                             \qquad\texttt{cvx\_hull}(\zbhat_{B, k}, \zb_k)
                           \end{array}}}\quad
                         \left( \mub_k^T \zbhat_{B,k}\ -\ \mub_{k+1}^T W_{k+1} \zb_{k} \right) \\
\end{equation}
Focusing on the second term that contains the minimization over the convex hull,
we will obtain a lower bound.
It is important, at this stage, to recall that, as seen in section \ref{sec:innerminpk}, the minimum of the second term can either be one of the three vertices of the triangle in Figure~\ref{fig:relucvx} (ambiguous ReLU), the $\zbhat_{B, k}=\zb_{k}$ line (passing ReLU), or the $(\zbhat_{B, k}=\mathbf{0},\zb_{k}=\mathbf{0})$ triangle vertex (blocking ReLU). We will write $(\zbhat_{B, k}, \zb_{k}) \in \texttt{ReLU\_sol}(\zbhat_{B, k}, \zb_k,\lb_k, \ub_k)$.

We can add a term $\lambdab_k^T(\sig{\zbhat_k} -
\sig{\zbhat_k}) = 0$ and obtain:
\begin{equation}
  \begin{split}
& \min_{\begin{array}{c}
        \zbhat_{B, k} \in [\lb_k, \ub_k]\\
        \mathclap{\texttt{cvx\_hull}(\zbhat_{B, k}, \zb_k)}
      \end{array}}
    \left( \mub_k^T \zbhat_{B,k}\ -\ \mub_{k+1}^T W_{k+1} \zb_{k} \right)\\
    = & \min_{\begin{array}{c}
    	\zbhat_{B, k} \in [\lb_k, \ub_k]\\
    	\mathclap{\texttt{ReLU\_sol}(\zbhat_{B, k}, \zb_k,\lb_k, \ub_k)}
    	\end{array}}
    \left( \mub_k^T \zbhat_{B,k}\ -\ \mub_{k+1}^T W_{k+1} \zb_{k} \right)\\
    = & \min_{\begin{array}{c}
        \zbhat_{B, k} \in [\lb_k, \ub_k]\\
        \mathclap{\texttt{ReLU\_sol}(\zbhat_{B, k}, \zb_k,\lb_k, \ub_k)}
      \end{array}}
    \left( \mub_k^T \zbhat_{B,k}\ -\ \mub_{k+1}^T W_{k+1} \zb_{k} + \lambdab_k^T(\sig{\zbhat_{B,k}} - \sig{\zbhat_{B,k}})\right)\\
    \geq & \min_{\begin{array}{c}
    	\zbhat_{B, k} \in [\lb_k, \ub_k]\\
    	\mathclap{\texttt{ReLU\_sol}(\zbhat_{B, k}, \zb_k,\lb_k, \ub_k)}
    	\end{array}} \left( \mub_k^T \zbhat_{B, k} - \lambdab_k^T\sig{\zbhat_{B,k}}\right)
    +\min_{\begin{array}{c}
        \zbhat_{B, k} \in [\lb_k, \ub_k]\\
        \mathclap{\texttt{ReLU\_sol}(\zbhat_{B, k}, \zb_k,\lb_k, \ub_k)}
      \end{array}} \left(- \mub_{k+1}^TW_{k+1}\zb_k  + \lambdab_k^T \sig{\zbhat_{B,k}}\right)\\
    \geq & \min_{\zbhat_{B,k} \in [\lb_k, \ub_k]} \left( \mub_k^T \zbhat_{B, k} - \lambdab_k^T\sig{\zbhat_{B,k}}\right)
    + \min_{\zb_k \in [\sig{\lb_k}, \sig{\ub_k}]} \left( -\mub_{k+1}^TW_{k+1}\zb_k  + \lambdab_k^T \zb_k\right)
  \end{split}
  \label{eq:lambpos_lb}
\end{equation}
Equality between the second line and the third comes from the fact that we
are adding a term equal to zero. The inequality between the third line and the
fourth is due to the fact that the sum of minimum is going to be lower than the
minimum of the sum. 
For what concerns obtaining the final line, the first term comes from projecting $\zb_{k}$ out of the feasible domain and taking the convex hull of the resulting domain. We need to look more carefully at the second term. Plugging in the ReLU formula:
\begin{equation*}
	\begin{split}
	&\min_{\begin{array}{c}
		\zbhat_{B, k} \in [\lb_k, \ub_k]\\
		\mathclap{\texttt{ReLU\_sol}(\zbhat_{B, k}, \zb_k,\lb_k, \ub_k)}
		\end{array}} \left(- \mub_{k+1}^TW_{k+1}\zb_k  + \lambdab_k^T \max\{0, \zbhat_{B,k}\}\right) \\
	= &\min_{\begin{array}{c}
		\zbhat_{B, k} \in [\sigma(\lb_k), \sigma(\ub_k)]\\
		\mathclap{\texttt{ReLU\_sol}(\zbhat_{B, k}, \zb_k,\lb_k, \ub_k)}
		\end{array}} \left(- \mub_{k+1}^TW_{k+1}\zb_k  + \lambdab_k^T \zbhat_{B,k}\right) \\
	\geq &\min_{\zb_{k} \in [\sigma(\lb_k), \sigma(\ub_k)]} \left(- \mub_{k+1}^TW_{k+1}\zb_k  + \lambdab_k^T \zb_{k}\right),
	\end{split}
\end{equation*}
as (keeping in mind the shape of $\texttt{ReLU\_sol}$ and for the purposes of this specific problem) excluding the negative part of the $\zbhat_{B,k}$ domain does not alter the minimal value. 
The final line then follows by observing that forcing $\zbhat_{B, k}=\zb$ is a convex relaxation of the positive part of $\texttt{ReLU\_sol}$.



Summing up the lower bounds for all the terms in \eqref{eq:dual_decomp_vs_dj},
as given by equations \eqref{eq:pb0_lb} and \eqref{eq:lambpos_lb},
we obtain the formulation of Problem~\eqref{eq:dj_dual_decomp}.
We conclude that the bound obtained by \citet{Dvijotham2018} is necessarily
no larger than the bound derived using our dual. Given that we are computing lower
bounds, this means that their bound is looser.

\end{proof}


	\section{Initialisation of the dual}
\label{sec:WK_init}
We will now show that the solution generated by the method of \citet{Wong2018}
can be used to provide initialization to our dual~\eqref{eq:nonprox_problem},
even though both duals were derived differently. Theirs is derived from the
Lagrangian dual of the Planet~\cite{Ehlers2017} relaxation, while ours come from
Lagrangian decomposition. However, we will show that a solution that can be
extracted from theirs lead to the exact same bound they generate.

As a reminder, we reproduce results proven in \citet{Wong2018}. In order to
obtain the bound, the problem they solved is:
\begin{equation}
  \begin{split}
    \min_{\zbhat_n} \quad &\mbf{c}^T \zbhat_n\\
    \text{such that } \quad&\zbhat_n \in \tilde{\mathcal{Z}}_\epsilon(x),
  \end{split}
\end{equation}
When $\mbf{c}$ is just an identity matrix (so we optimise the bounds of each
neuron), we recover the problem ~\eqref{eq:planet-relax}.
$\tilde{\mathcal{Z}}_\epsilon(x)$ correspond to the contraints
\eqref{eq:planet-inibounds} to \eqref{eq:planet-linconstr}.

Based on the {\bf Theorem 1}, and the equations (8) to (10)~\citep{Wong2018},
the dual problem that they derive has an objective function of\footnote{We
  adjust the indexing slightly because Wong et al. note the input $\zb_1$, while
  we refer to it as $\zb_0$}:
\begin{equation}
  -\sum_{i=1}^{n-1} \nub_{i}^T \mbf{b}_i - \xb^T \hat{\nub}_0 - \epsilon \| \hat{\nub}_0 \|_1 + \sum_{i=1}^{n-1}\sum_{j \in \mathcal{I}_i} l_{i,j} \left[ \nu_{i,j}\right]_{+}
  \label{eq:kw_obj}
\end{equation}
with a solution given by
\begin{equation}
  \begin{split}
    \nub_n &= - \mbf{c} \\
    \hat{\nub}_i &= - W_{i+1}^TD_{i+2}W_{i+2}^T \dots D_{n}W_{n}^T\mbf{c}\\
    \nub_i &= D_i \hat{\nub}_i
  \end{split}
\end{equation}
where $D_i$ is a diagonal matrix with entries
\begin{equation}
  (D_i)_{jj} =
  \begin{cases}
    0 & j \in \mathcal{I}_i^-\\
    1 & j \in \mathcal{I}_i^+\\
    \frac{u_{i,j}}{u_{i,j}-l_{i,j}} & j \in \mathcal{I}_i
  \end{cases},
\end{equation}
and $\mathcal{I}_i^-, \mathcal{I}_i^+, \mathcal{I}_i$ correspond respectively to
blocked ReLUs (with always negative inputs), passing ReLUs (with always positive
inputs) and ambiguous ReLUs.

We will now prove that, by taking as our solution of~\eqref{eq:nonprox_problem}
\begin{equation}
  \lambdabhat_k = \nub_{k},
  \label{eq:kw_to_ours_dual}
\end{equation}
we obtain exactly the same bound.

The problem we're solving is~\eqref{eq:nonprox_problem}, so, given a choice of
$\lambdabhat$, the bound that we generate is:
\begin{equation}
  \begin{split}
    \text{min }_{\zb, \zbhat}\ & \zbhat_{A, n}
    + \hspace{-8pt}\sum_{k=1..n-1}\hspace{-8pt} \lambdabhat_k^T \left( \zbhat_{B, k} - \zbhat_{A, k} \right)\\
    \text{s.t. }\quad & \mathcal{P}_0(\zb_0, \zbhat_{A,1}); \qquad \mathcal{P}_k(\zbhat_{B, k}, \zbhat_{A, k+1})\quad \text{for } k \in \left[1..n-1\right].\\
  \end{split}
\end{equation}
which can be decomposed into several subproblems:
\begin{equation}
  \sum_{k=1}^{n-1}\left(
    \min_{\begin{array}{c}\zbhat_{B, k}, \zbhat_{A, k+1} \\ \in \mathcal{P}_k\left( \cdot, \cdot \right)\end{array}}
    \lambdabhat_k^T \zbhat_{B,k}\ -\ \lambdabhat_{k+1}^T \zbhat_{A, k+1}
  \right)
  + \min_{\begin{array}{c}\zb_0, \zbhat_{A,1}\\\in \mathcal{P}_0\left( \cdot, \cdot \right)\end{array}} -\lambdabhat_1^T \zbhat_{A, 1}
  \label{eq:dual_decomp_for_kw}
\end{equation}
where we take the convention consistent with~\eqref{eq:kw_to_ours_dual} that
\mbox{$\lambdabhat_n = \nub_n = -\mathbf{c} = - I$}.

Let's start by evaluating the problem over $\mathcal{P}_0$, having replaced
$\lambdabhat_1$ by $\nub_1$ in accordance with~\eqref{eq:kw_to_ours_dual}:
\begin{equation}
  \begin{split}
    \min_{\zb_0, \zbhat_{A, 1}}\quad& - \nub_1^T \zbhat_{A, 1}\\
    \text{such that }\quad&  \zbhat_{1} = W_{1} \zb_{0} + \mathbf{b}_{1}\\
    & \zb_{0} \in \mathcal{C},
  \end{split}
\end{equation}
where in the context of \citep{Wong2018}, $\mathcal{C}$ is defined by the bounds
\mbox{$\lb_0 = x - \epsilon$} and \mbox{$\ub_0 = x + \epsilon$}. We already
explained how to solve this subproblem in the context of our Frank Wolfe
optimisation, see Equation~\eqref{eq:box_bounds_sol}). Plugging in the solution
into the objective function gives us:
\begin{equation}
  \begin{split}
  \min_{\begin{array}{c}\zb_0, \zbhat_{A,1}\\\in \mathcal{P}_0\left( \cdot, \cdot \right)\end{array}}
  -\lambdabhat_1^T \zbhat_{A, 1}
  =& -\nub_1^TW_1\left( \mathbb{1}_{\nub_1^T W_1> 0} \cdot \lb_{0} +\mathbb{1}_{ \nub_1^T W_1 \leq 0} \cdot \ub_0\right)
   -\nub_1^T b_1\\
   =&-\hat{\nub}_0^T\left( \mathbb{1}_{\hat{\nub}_0> 0} \cdot \lb_{0} +\mathbb{1}_{ \hat{\nub}_0 \leq 0} \cdot \ub_0\right)
   -\nub_1^T b_1\\
   =&- \hat{\nub}_0^T \xb + \epsilon\hat{\nub}_0^T \mathbb{1}_{\hat{\nub}_0> 0} - \epsilon \hat{\nub}_0^T \mathbb{1}_{\hat{\nub}_0< 0}- \nub_1^T b_1\\
   =& -\hat{\nub}_0^T \xb + \epsilon \| \hat{\nub}_0 \|_1 -\nub_1^T \mbf{b}_1
  \end{split}
\end{equation}

We will now evaluate the values of the problem over $\mathcal{P}_k$. Once again,
we replace the $\lambdabhat$ with the appropriate values of $\nub$ in accordance
to \eqref{eq:kw_to_ours_dual}:
\begin{equation}
  \begin{split}
    \min_{\zbhat_{B, k}, \zbhat_{A, k+1}}& \quad
    \nub_{k}^T \zbhat_{B, k} + \nub_{k}^T \zbhat_{A, k+1}\\
    \text{s.t } & \quad \lb_{k} \leq \zbhat_{B, k} \leq \ub_{k}\\
    & \quad \texttt{cvx\_hull}_\sigma\left( \zbhat_{B,k}, \zb_k, \lb_k, \ub_k\right)\\
    & \quad \zbhat_{A, k+1} = W_{k+1} \zb_{k} + \mathbf{b}_{k+1},
  \end{split}
\end{equation}

We can rephrase the problem such that it decomposes completely over the ReLUs,
by merging the last equality constraint into the objective function:
\begin{equation}
  \begin{split}
    \min_{\zbhat_{B, k}, \zb_{k}}& \quad
    \nub_{k}^T \zbhat_{B, k} - \nub_{k+1}^T W_{k+1} \zb_{k} - \nub_{k+1}^T \mathbf{b}_{k+1}\\
    \text{s.t } & \quad \lb_{k} \leq \zbhat_{B, k} \leq \ub_{k}\\
    & \quad \texttt{cvx\_hull}_\sigma\left( \zbhat_{B,k}, \zb_k, \lb_k, \ub_k\right)\\
  \end{split}
\label{eq:pk_problem_kw}
\end{equation}
Note that \mbox{$\nub_{k+1}^TW_{k+1} = \hat{\nub}_{k}^T$}. We will now omit the
term $-\nub_{k+1}^T \mbf{b}_{k+1}$ which doesn't depend on the optimisation
variable anymore and will therefore be directly passed down to the value of the
bound.

For the rest of the problem, we will distinguish the different cases of ReLU,
$\mathcal{I}^+, \mathcal{I}^-$ and $\mathcal{I}$.

In the case of a ReLU $j \in \mathcal{I}^+$, we have
\mbox{$\texttt{cvx\_hull}_\sigma\left( \dots \right) \quad \equiv \quad
  \zbhat_{B, k}[j] = \zb_k[j]$}, so the replacement of $\zb_k[j]$ makes the
objective function term for $j$ becomes $\left( \nub_{k}[j] -
  \hat{\nub}_{k}[j]\right)\zbhat_{B,k}[j]$. Given that $j \in \mathcal{I}^+$,
this means that the corresponding element in $D_{k}$ is 1, so $\nub_{k}[j] =
\hat{\nub}_{k}[j]$, which means that the term in the objective function is
zero.

In the case of a ReLU $j \in \mathcal{I}^-$, we have
\mbox{$\texttt{cvx\_hull}_\sigma\left( \dots \right) \quad \equiv \quad \zb_k[j]
  = 0$}, so the replacement of $\zb_k[j]$ makes the objective function term for
$j$ becomes $\nub_{k}[j]\zbhat_{B,k}[j]$. Given that $j \in
\mathcal{I}^-$, this means that the corresponding element in $D_{k}$ is 0, so
$\nub_{k}[j] = 0$. This means that the term in the objective function is
zero.

The case for a ReLU $j \in \mathcal{I}$ is more complex. We apply the same
strategy of picking the appropriate inequalities in $\texttt{cvx\_hull}$ as we
did for obtaining Equation~\eqref{eq:ambrelu_linopt}, which leads us to:
\begin{equation}
  \begin{split}
    \min_{\zbhat_{B, k}}\quad & (\nub_{k}[j] - [\hat{\nub}_{k}[j]]_+ \frac{u_k[j]}{u_k[j]-l_k[j]})\zb_{B,k}[j] \\
     & - [\hat{\nub}_{k}[j]]_- \max(\zb_{B,k}, 0)\\
     & + [\hat{\nub}_{k}[j]]_+ \frac{u_k[j]}{u_k[j]-l_k[j]}l_k[j]\\
    \text{s.t } \quad &\lb_{k} \leq \zbhat_{B, k} \leq \ub_{k}\\
  \end{split}
\end{equation}
Once again, we can recognize that the coefficient in front of $\zb_{B,k}[j]$ (in
the first line of the objective function) is equal to 0 by construction of the
values of $\nub$ and $\hat{\nub}$. The coefficient in front of $\max(\zb_{B,k},
0)$ in the second line is necessarily positive. Therefore, the minimum value for
this problem will necessarily be the constant term in the third line (reachable
for $\zb_{B, k} = 0$ which is feasible due to $j$ being in $\mathcal{I}$).
The value of this constant term is:
\begin{equation}
    \left[ \hat{\nub}_{k}[j]\right]_+ \frac{u_k[j]}{u_k[j]-l_k[j]}l_k[j] = \left[ \nub_{k}[j]\right]_+ l_k[j]
\end{equation}

Regrouping the different cases and the constant term of
problem~\eqref{eq:pk_problem_kw} that we ignored, we find that the value of the
term corresponding to the $\mathcal{P}_k$ subproblem is:
\begin{equation}
  -\nub_{k+1}^T \mbf{b}_{k+1} + \sum_{j \in \mathcal{I}}\left[\nub_{k}[j]\right]_+ l_k[j]
\end{equation}

Plugging all the optimisation result into Equation~\eqref{eq:dual_decomp_for_kw}, we
obtain:
\begin{equation}
  \begin{split}
  \sum_{k=1}^{n-1}&\left(
    \min_{\mathcal{P}_k}
    \lambdabhat_k^T \zbhat_{B,k}-\lambdabhat_{k+1}^T \zbhat_{A, k+1} \right) + \min_{\mathcal{P}_0} -\lambdabhat_1^T \zbhat_{A, 1}\\
  & = \sum_{k=1}^{n-1}\left(
    -\nub_{k+1}^T \mbf{b}_{k+1} + \sum_{j \in \mathcal{I}}\left[\nub_{k}[j]\right]_+ l_k[j]\right)
  -\hat{\nub}_0^T \xb + \epsilon \| \hat{\nub}_0 \|_1 -\nub_1^T \mbf{b}_1\\
  & = -\sum_{k=1}^{n-1} \nub_{k}^T\mbf{b}_{k} - \hat{\nub}_0^T \xb - \epsilon \|\hat{\nub}_0\|_1
  + \sum_{k=1}^{n-1}\sum_{j \in \mathcal{I}}\left[\nub_{k}[j]\right]_+ l_k[j],
  \end{split}
\end{equation}
which is exactly the value given by \citet{Wong2018} in Equation~\eqref{eq:kw_obj}.


	\newpage

\section{Supplementary experiments}
We now complement the results presented in the main paper and provide additional information on the experimental setting. We start from incomplete verification ($\S$ \ref{sec:experiments-incomplete-appendix}) and then move on to complete verification  ($\S$ \ref{sec:experiments-complete-appendix}).

\subsection{Incomplete Verification} \label{sec:experiments-incomplete-appendix}

Figure~\ref{fig:madry-add-pointwise} presents additional pointwise comparisons for the distributional results showed in Figure~\ref{fig:madry8}. The comparison between Dec-DSG+ and DSG+ shows that the inequality in Theorem~\ref{theorem} is strict for most of the obtained dual solutions, but operating directly on problem \eqref{eq:nonprox_problem} is preferable in the vast majority of the cases. For completeness, we also include a comparison between WK and Interval Propagation (IP), showing that the latter yields significantly worse bounds on most images, reason for which we omitted it in the main body of the paper.

\begin{figure*}[h!]
	\begin{subfigure}{.495\textwidth}
		\centering
		\includegraphics[width=\textwidth]{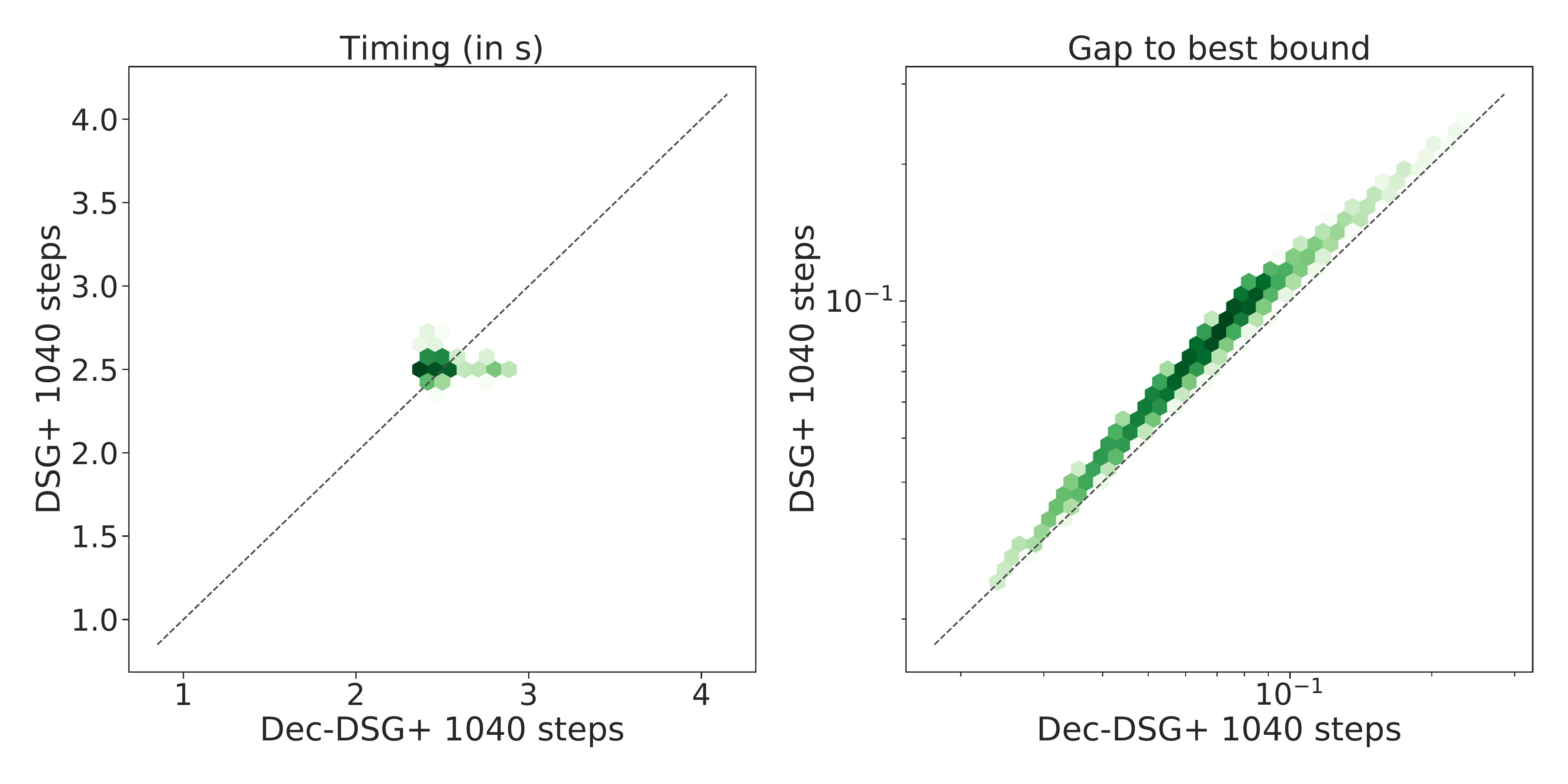}
	\end{subfigure}
	\begin{subfigure}{.495\textwidth}
		\centering
		\includegraphics[width=\textwidth]{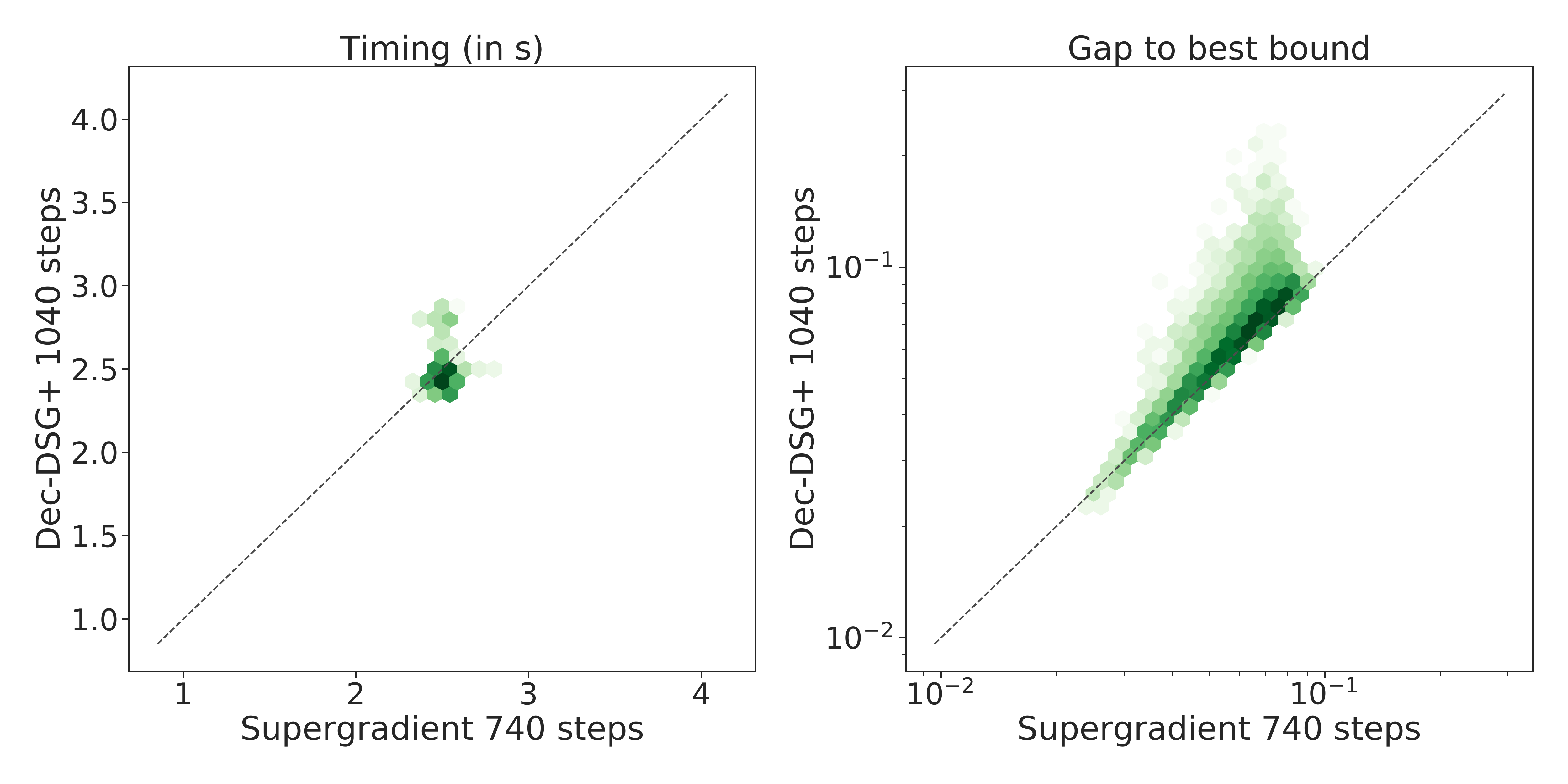}
	\end{subfigure}
	\\
	\begin{subfigure}{.495\textwidth}
		\centering
		\includegraphics[width=\textwidth]{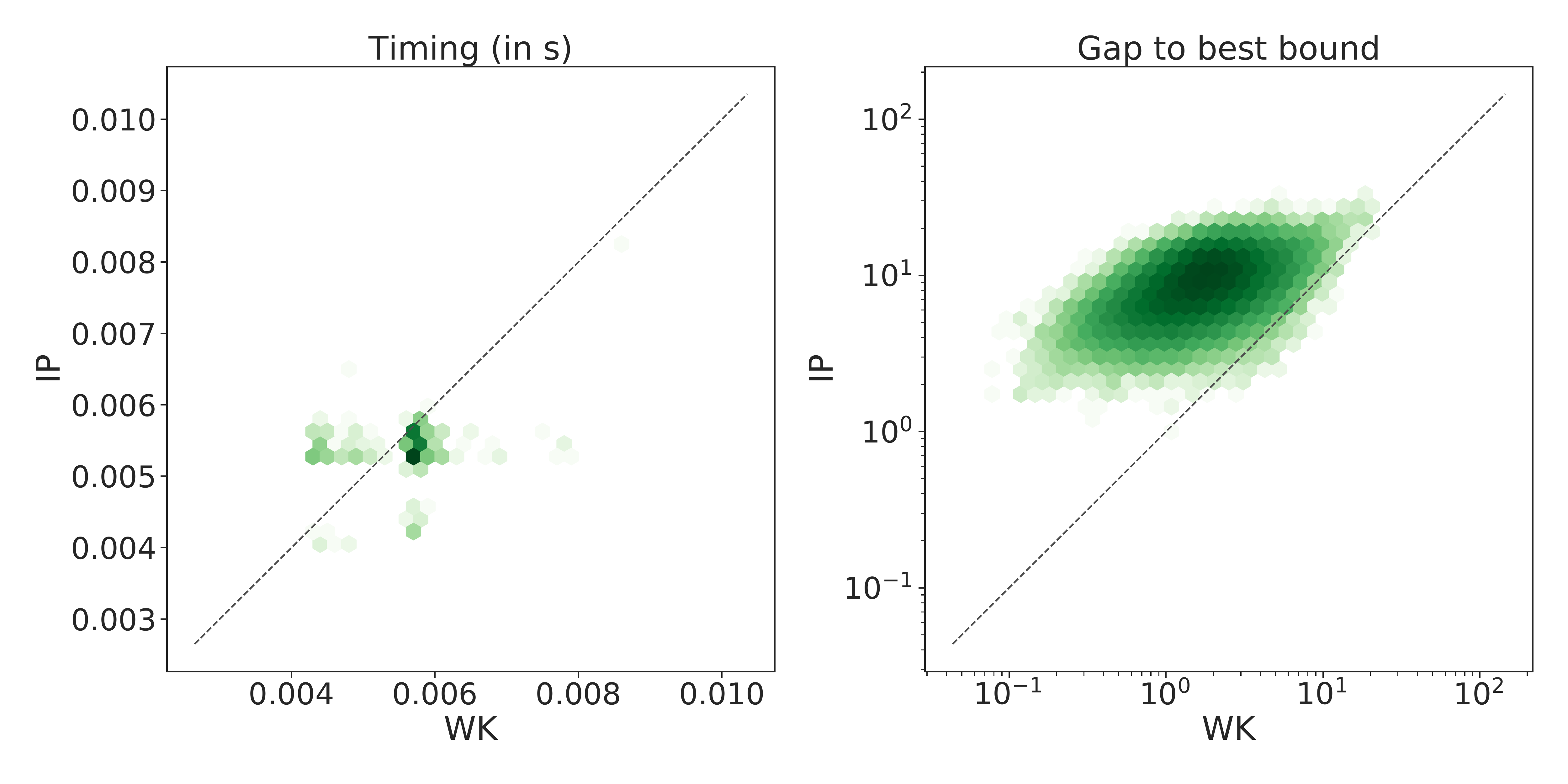}
	\end{subfigure}
		\hspace*{20pt}
	\begin{minipage}{.4\textwidth}
		\caption{\label{fig:madry-add-pointwise}Additional pointwise comparison for a subset of the methods in Figure \ref{fig:madry8}, along with the the omitted Interval Propagation (IP). Each datapoint corresponds to a CIFAR image, darker colour shades mean higher point density. The dotted line corresponds to the equality and in both graphs, lower is better along both axes.}
	\end{minipage}
\end{figure*}

\begin{figure*}[h!]
	\centering
	\includegraphics[width=1\textwidth]{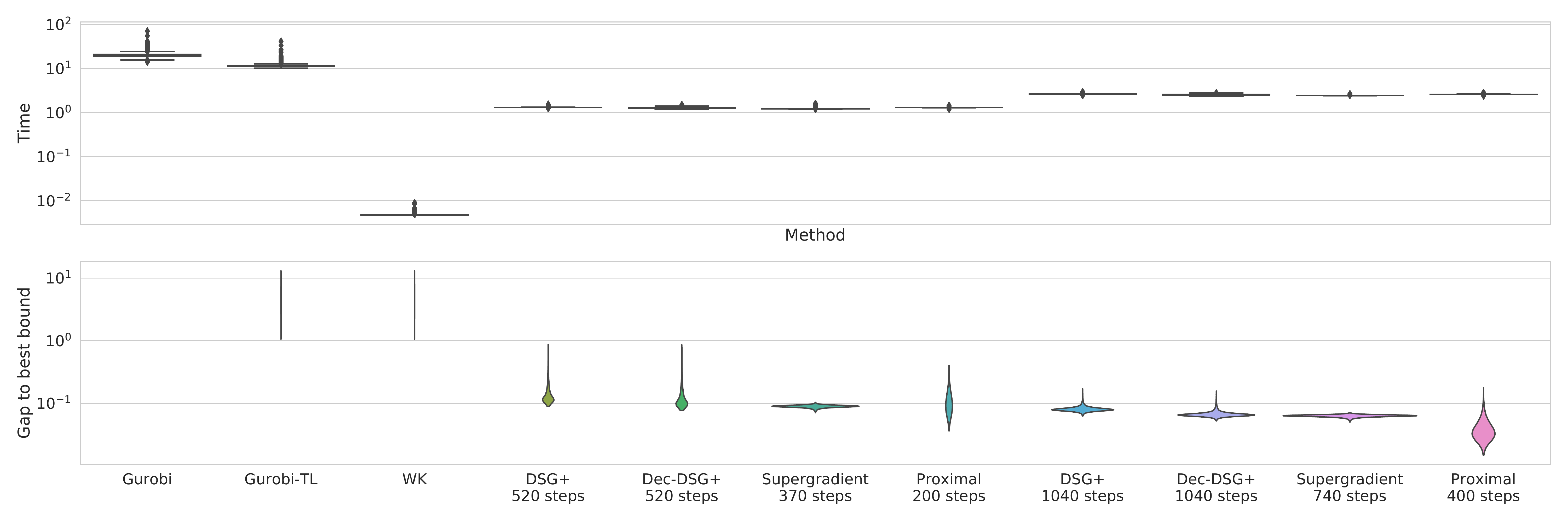}
	\caption{\label{fig:sgd8} Comparison of the distribution of runtime
		and gap to optimality on a normally trained network. In both
		cases, lower is better. The width at a given value represents the
		proportion of problems for which this is the result. Note that Gurobi
		always return the optimal solution so doesn't appear on the optimality
		gap, but is always the highest runtime.}
		\vspace{20pt}
\end{figure*}

Figures \ref{fig:sgd8}, \ref{fig:sgd-pointwise}, \ref{fig:sgd-add-pointwise} show experiments (see section \ref{sec:experiments-incomplete}) for a network trained using standard stochastic gradient descent and cross entropy, with no robustness objective. We employ $\epsilon_{\text{verif}} = 5/255$. Most observations done in the main paper for the adversarially trained network (Figures \ref{fig:madry8} and \ref{fig:madry8-pointwise}) still hold true: in particular, the advantage of the Lagrangian Decomposition based dual compared to the dual by \citet{Dvijotham2018} is even more marked than in Figure \ref{fig:madry8-pointwise}. In fact, for a subset of the properties, DSG+ has returns rather loose bounds even after $1040$ iterations. 
The proximal algorithm returns better bounds than the supergradient method in this case as well, even though the larger support of the optimality gap distribution and the larger runtime might reduce the advantages of the proximal algorithm.

\begin{figure*}[h!]
	\centering
	\begin{subfigure}{.495\textwidth}
		\centering
		\includegraphics[width=\textwidth]{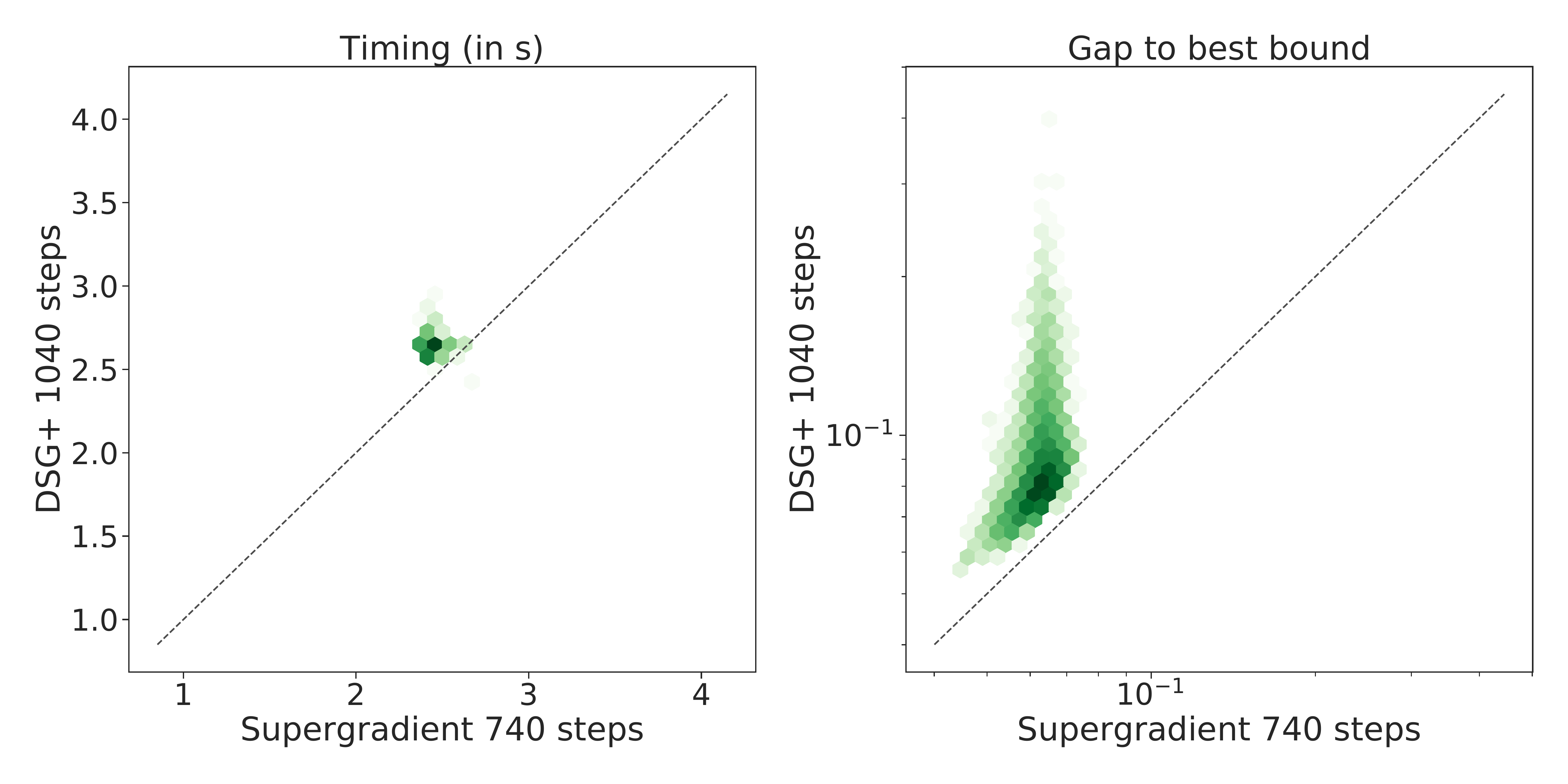}
	\end{subfigure}
	\begin{subfigure}{.495\textwidth}
		\centering
		\includegraphics[width=\textwidth]{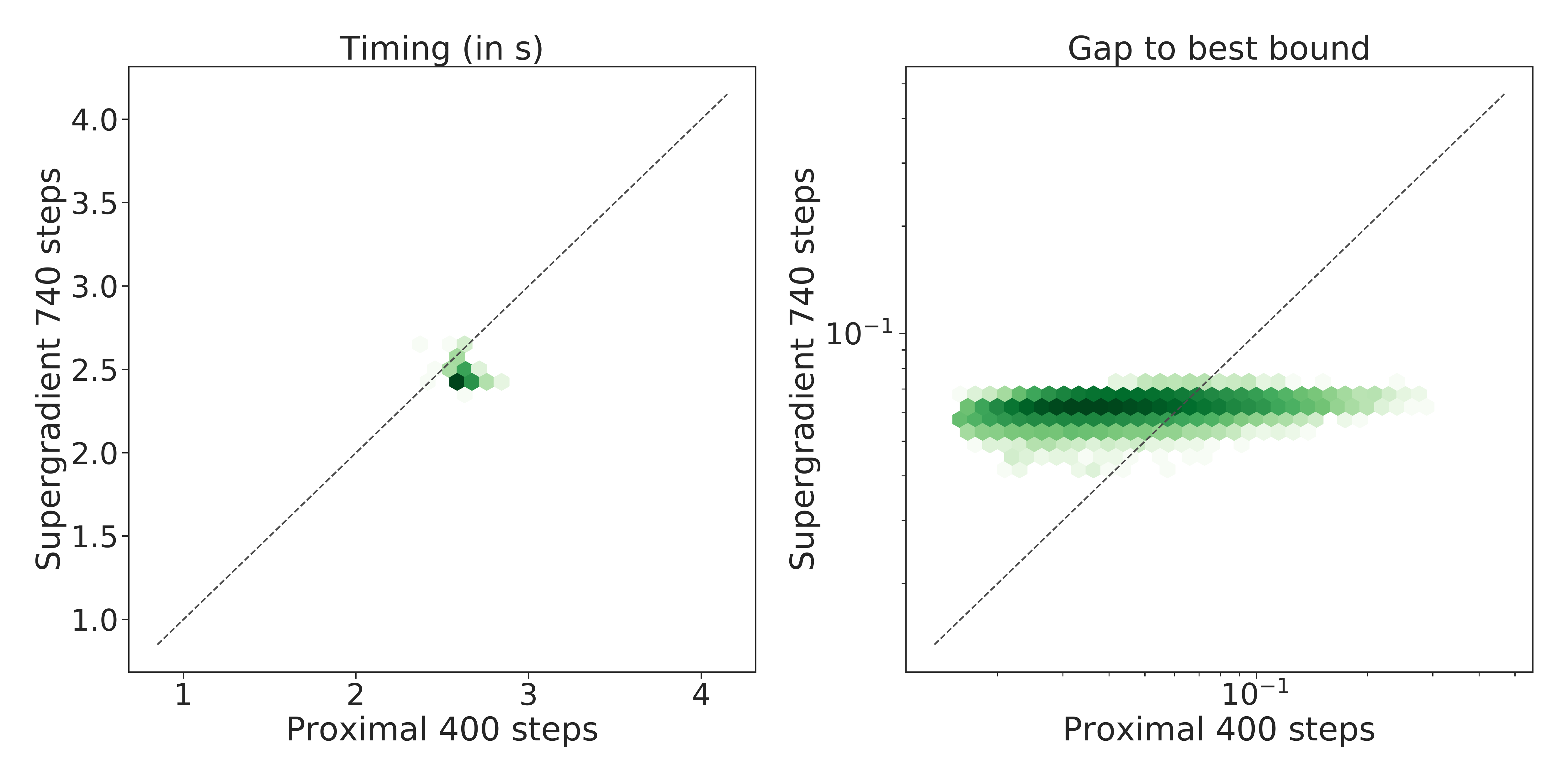}
	\end{subfigure}
	\caption{Pointwise comparison for a subset of the methods in Figure \ref{fig:sgd8}. Each datapoint corresponds to a CIFAR image, darker colour shades mean higher point density. The dotted line corresponds to the equality and in both graphs, lower is better along both axes.}
	\label{fig:sgd-pointwise}
\end{figure*}

\begin{figure*}[h!]
	\begin{subfigure}{.495\textwidth}
		\centering
		\includegraphics[width=\textwidth]{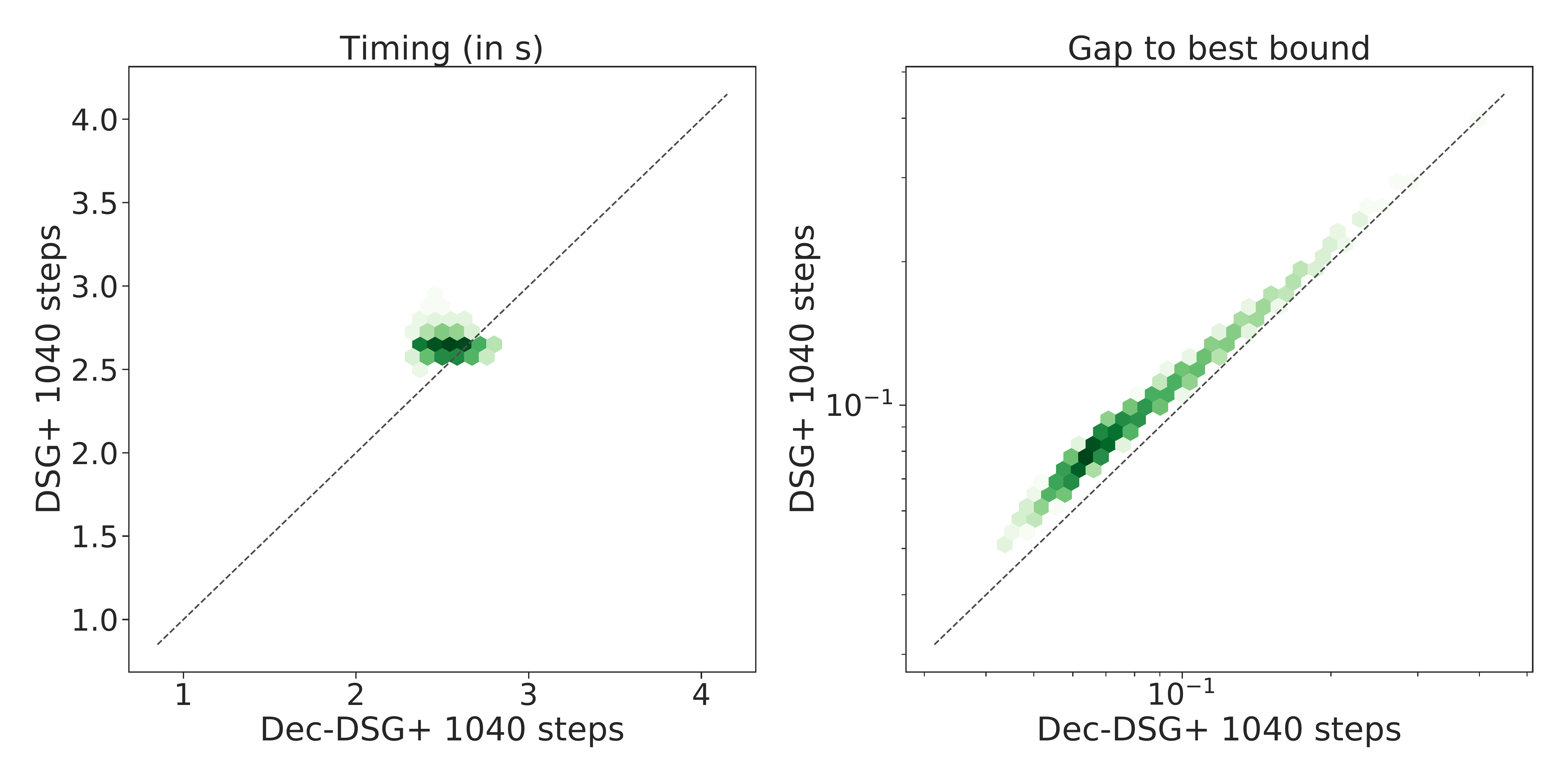}
	\end{subfigure}
	\begin{subfigure}{.495\textwidth}
		\centering
		\includegraphics[width=\textwidth]{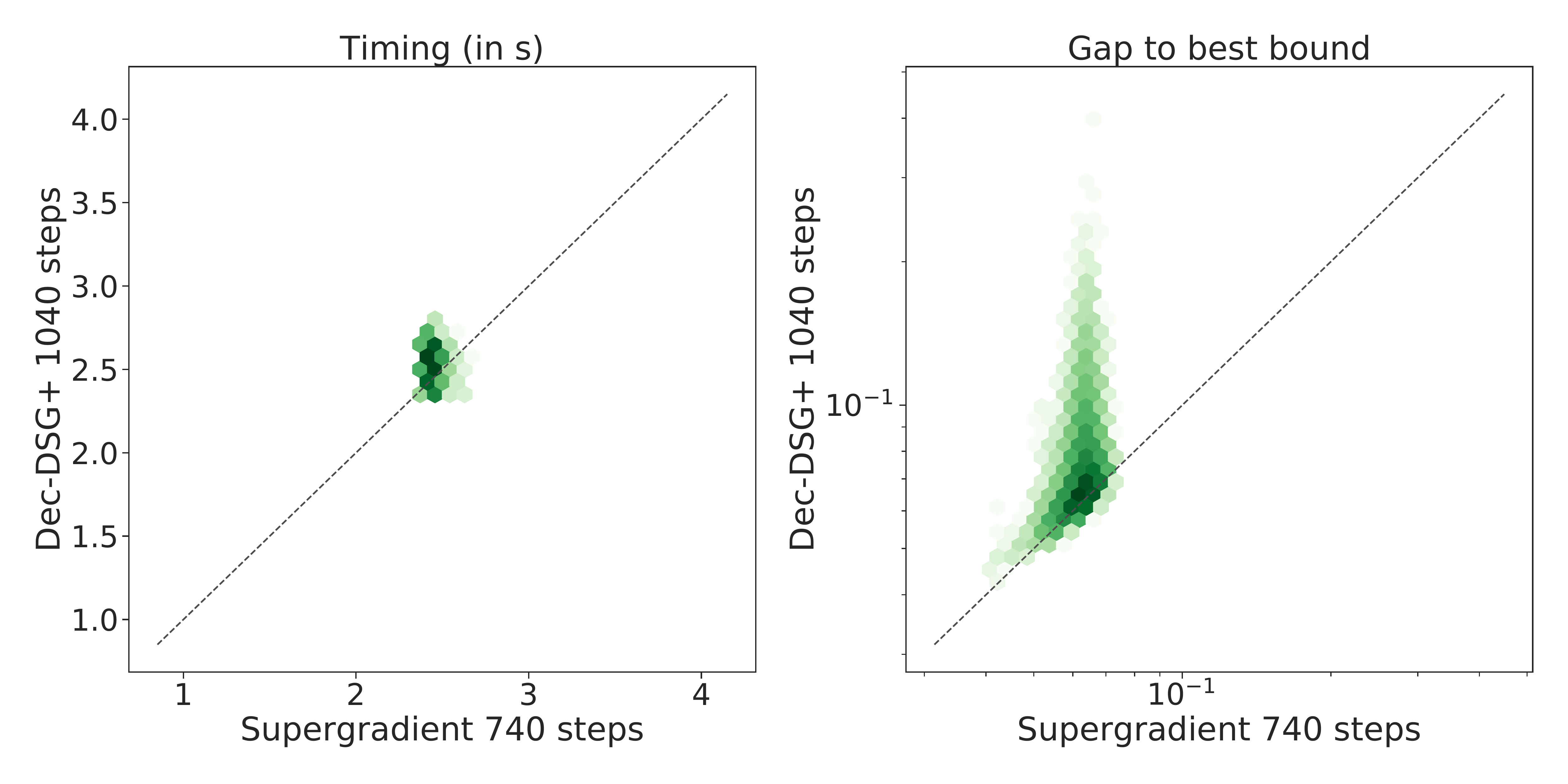}
	\end{subfigure}
	\\
	\begin{subfigure}{.495\textwidth}
		\centering
		\includegraphics[width=\textwidth]{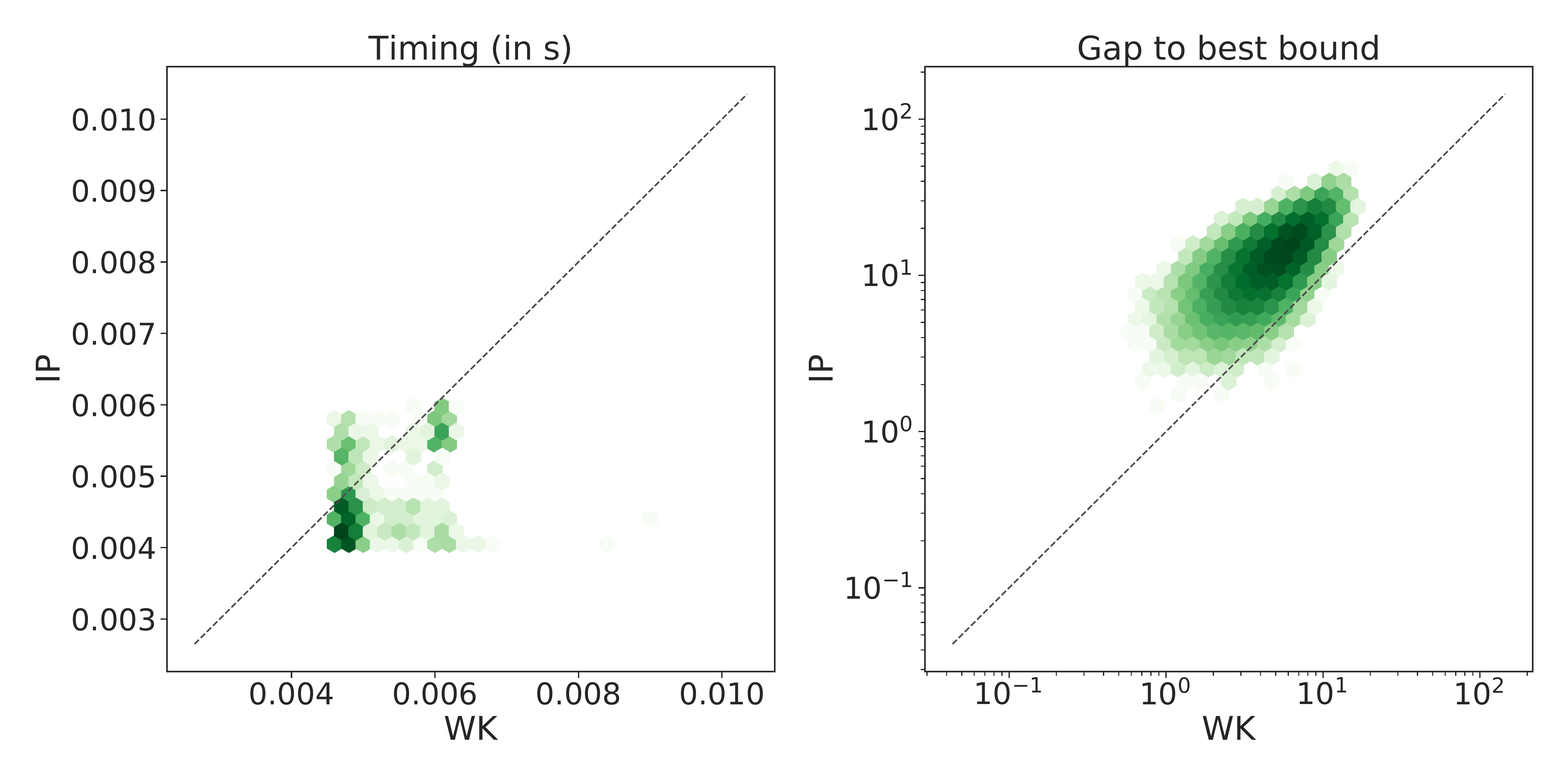}
	\end{subfigure}
	\hspace*{20pt}
	\begin{minipage}{.4\textwidth}
		\caption{\label{fig:sgd-add-pointwise} Additional pointwise comparison for a subset of the methods in Figure \ref{fig:sgd8}, along with the the omitted Interval Propagation (IP). Each datapoint corresponds to a CIFAR image, darker colour shades mean higher point density. The dotted line corresponds to the equality and in both graphs, lower is better along both axes.}
	\end{minipage}
\end{figure*}

\vspace{20pt}
\subsubsection{Network Architecture}
\label{sec:network_arch}

We now outline the network architecture for both the adversarially trained and the normally trained network experiments.

\begin{table}[h!]
	\vspace{20pt}
	\centering
	\begin{tabular}{|c|}
		\toprule
		Conv2D(channels=16, kernel\_size=4, stride=2)\\
		\midrule
		Conv2D(channels=32, kernel\_size=4, stride=2)\\
		\midrule
		Linear(channels=100)\\
		\midrule
		Linear(channels=10)\\
		\bottomrule
	\end{tabular}
	\caption{Network architecture for incomplete verification, from \citet{Wong2018}. Each layer but the last is followed by a ReLU activation function.}
	\label{tab:network_arch}
		\vspace{10pt}
\end{table}

\subsection{Complete Verification} \label{sec:experiments-complete-appendix}

As done by \citet{Lu2020}, we provide a more in-depth analysis of the base model results presented in Figure \ref{fig:verification-appendix}. Depending on the time $t_i$ that the Gurobi baseline employed to verify them, the properties have been divided into an ``easy" ($t_i < 800$), ``medium" and a ``hard " ($t_i > 2400$) set. 
The relative performance of the iterative dual methods remains unvaried, with the initial gap between our two methods increasing with the difficulty of the problem. \vspace{5pt}

\sisetup{detect-weight=true,detect-inline-weight=math,detect-mode=true}
\begin{table}[h!]
	\centering
	\caption{\small For easy, medium and difficult level verification properties on the base model, we compare average solving time, average number of solved sub-problems (on the properties that did not time out) and the percentage of timed out properties.}
	\label{table:base-model}
	\scriptsize
	\setlength{\tabcolsep}{4pt}
	\aboverulesep = 0.1mm  
	\belowrulesep = 0.2mm  
	\begin{adjustbox}{center}
		\begin{tabular}{
				l
				S[table-format=4.6]
				S[table-format=4.6]
				S[table-format=4.6]
				S[table-format=4.6]
				S[table-format=4.6]
				S[table-format=4.6]
				S[table-format=4.6]
				S[table-format=4.6]
				S[table-format=4.6]
			}
			& \multicolumn{3}{ c }{Easy} & \multicolumn{3}{ c }{Medium} & \multicolumn{3}{ c }{Hard} \\
			\toprule
			
			\multicolumn{1}{ c }{Method} &
			\multicolumn{1}{ c }{time(s)} &
			\multicolumn{1}{ c }{sub-problems} &
			\multicolumn{1}{ c }{$\%$Timeout} &
			\multicolumn{1}{ c }{time(s)} &
			\multicolumn{1}{ c }{sub-problems} &
			\multicolumn{1}{ c }{$\%$Timeout} &
			\multicolumn{1}{ c }{time(s)} &
			\multicolumn{1}{ c }{sub-problems} &
			\multicolumn{1}{ c }{$\%$Timeout} \\
			
			\cmidrule(lr){1-1} \cmidrule(lr){2-4} \cmidrule(lr){5-7} \cmidrule(lr){8-10}
			
			\multicolumn{1}{ c }{\textsc{Gurobi-BaBSR}}
			&545.720
			&580.428
			&0.0
			
			&1370.395
			&1408.745
			&0.0
			
			&3127.995
			&2562.872
			&41.31 \\
			
			\multicolumn{1}{ c }{\textsc{MIPplanet}}
			& 1499.375
			& 
			&16.48
			
			&2240.980
			& 
			&42.94
			
			&2250.624
			& 
			&46.24 \\
			
			\multicolumn{1}{ c }{\textsc{DSG+ BaBSR}}
			&112.100
			&2030.362
			&0.85
			
			&177.805
			&4274.811
			&0.52
			
			&1222.632
			&12444.449
			&23.71 \\
			
			\multicolumn{1}{ c }{\textsc{Supergradient BaBSR}}
			& 85.004
			& 1662.000
			& 0.64
			
			& 136.853
			& 3772.679
			& 0.26
			
			& 1133.264
			& 12821.481
			&\B 21.36 \\
			
			\multicolumn{1}{ c }{\textsc{Proximal BaBSR}}
			&\B 73.606
			&\B 1377.666
			&\B 0.42
			
			&\B 129.537
			&\B 3269.535
			&\B 0.26
			
			&\B 1126.856
			&\B 12035.994
			& 21.83 \\
			
			\bottomrule
			
		\end{tabular}
	\end{adjustbox}
\end{table}

\begin{figure*}[h!]
	\centering
	\begin{subfigure}{.33\textwidth}
		\centering
		\includegraphics[width=\textwidth]{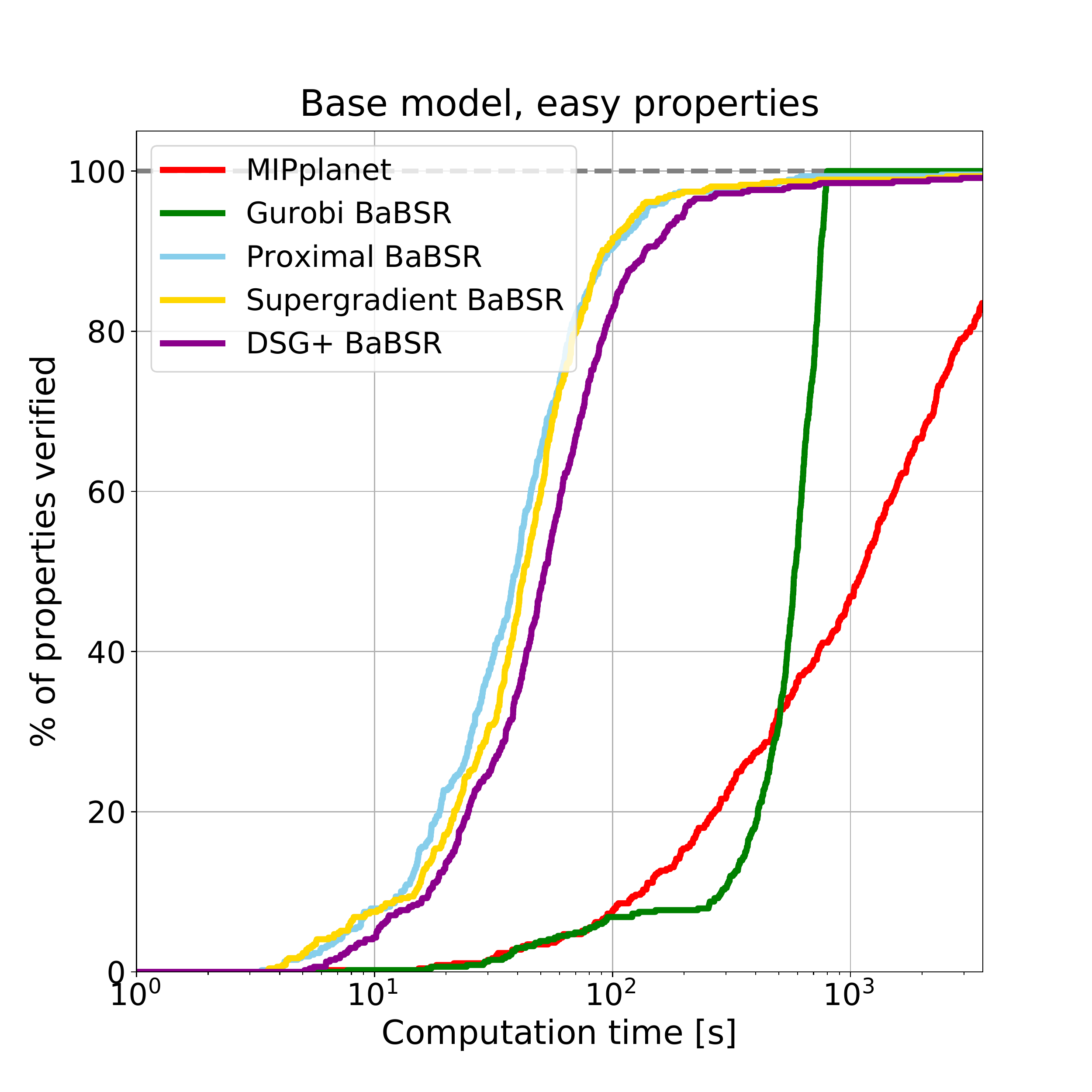}
	\end{subfigure}
	\begin{subfigure}{.33\textwidth}
		\centering
		\includegraphics[width=\textwidth]{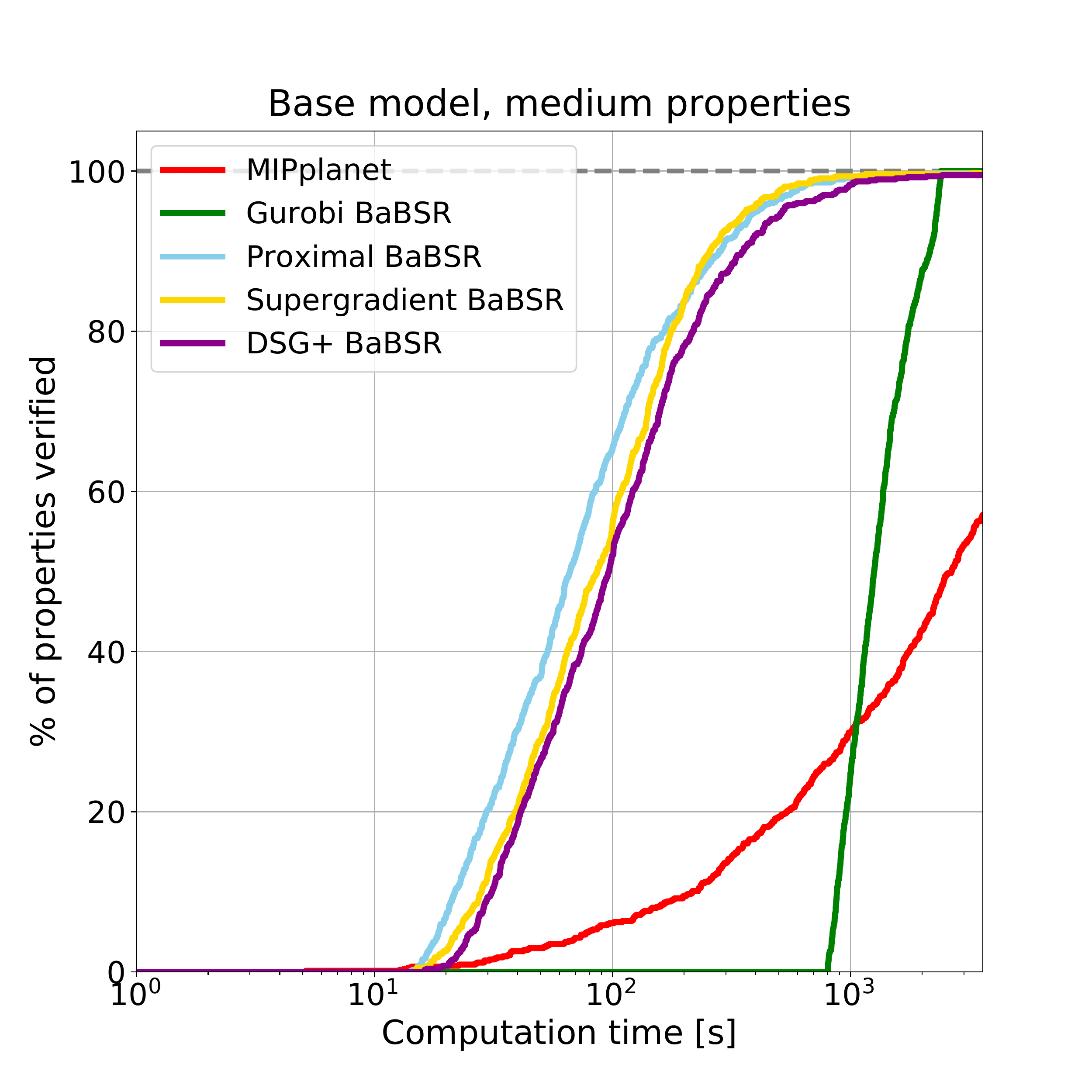}
	\end{subfigure}
	\begin{subfigure}{.33\textwidth}
		\centering
		\includegraphics[width=\textwidth]{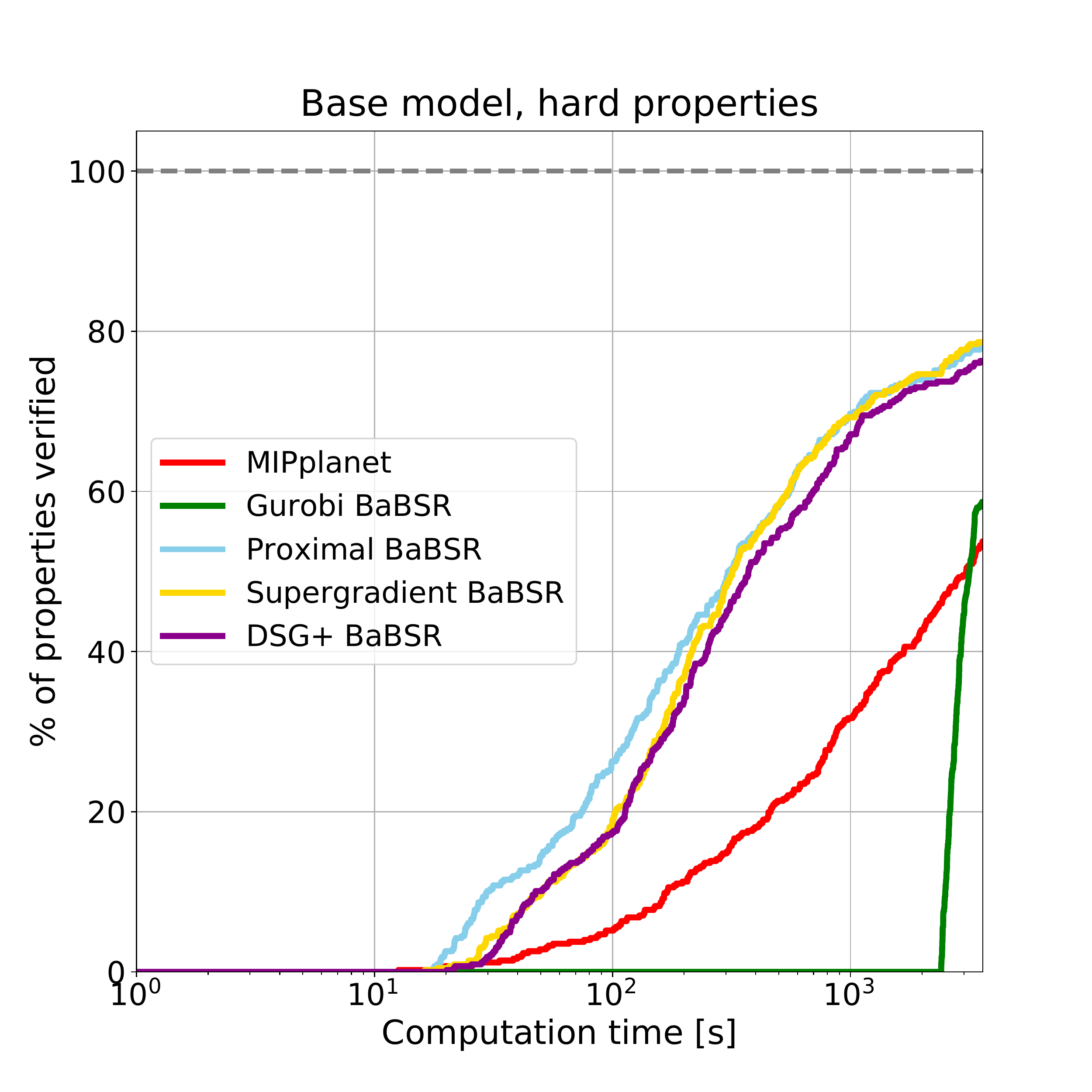}
	\end{subfigure}
	\caption{Cactus plots for the base model, with properties of varying difficulty. For each, we compare the bounding methods and complete verification algorithms by plotting the percentage of solved properties as a function of runtime.}
	\label{fig:verification-appendix}
\end{figure*}

\subsubsection{Network Architectures}
\label{sec:network_arch_verification}

Finally, we describe properties and network architecture for the dataset by \citet{Lu2020}.

\begin{table}[h!]
	\centering
	\small
	\begin{tabular}{|c|c|c|}
		\hline
		\textbf{Network Name}\TBstrut & \textbf{No. of Properties} \TBstrut & \textbf{Network Architecture} \TBstrut\\
		\hline
		\begin{tabular}{l}
			BASE \\ 
			Model
		\end{tabular} & \begin{tabular}{@{}c@{}}
			Easy: 467  \\
			Medium: 773 \\
			Hard: 426
		\end{tabular} & \begin{tabular}{@{}c@{}} 
			\footnotesize
			Conv2d(3,8,4, stride=2, padding=1) \Tstrut \\
			Conv2d(8,16,4, stride=2, padding=1)\\
			linear layer of 100 hidden units \\
			linear layer of 10 hidden units\\
			(Total ReLU activation units: 3172) \Bstrut
		\end{tabular}\\
		\hline
		WIDE & 300 & \begin{tabular}{@{}c@{}} 
			\footnotesize
			Conv2d(3,16,4, stride=2, padding=1) \Tstrut\\
			Conv2d(16,32,4, stride=2, padding=1)\\
			linear layer of 100 hidden units \\
			linear layer of 10 hidden units\\
			(Total ReLU activation units: 6244) \Bstrut
		\end{tabular}\\
		\hline
		DEEP & 250 & \begin{tabular}{@{}c@{}} 
			\footnotesize
			Conv2d(3,8,4, stride=2, padding=1) \Tstrut \\
			Conv2d(8,8,3, stride=1, padding=1) \\
			Conv2d(8,8,3, stride=1, padding=1) \\
			Conv2d(8,8,4, stride=2, padding=1)\\
			linear layer of 100 hidden units \\
			linear layer of 10 hidden units\\
			(Total ReLU activation units: 6756) \Bstrut
		\end{tabular}\\
		\hline
	\end{tabular}
	\caption{\label{tab:problem_size} For each complete verification experiment, the network architecture used and the number of verification properties tested, from the dataset by \citet{Lu2020}. Each layer but the last is followed by a ReLU activation function.}
\end{table}

}{}

\end{document}